\documentclass[11pt]{article}

\usepackage[preprint]{acl}

\usepackage{times}
\usepackage{latexsym}
\usepackage[T1]{fontenc}
\usepackage[utf8]{inputenc}
\usepackage{microtype}

\usepackage{amsmath}
\usepackage{amssymb}
\usepackage[inline,shortlabels]{enumitem}
\usepackage{graphicx}
\usepackage{booktabs}
\usepackage{multirow}
\usepackage{todonotes}
\usetikzlibrary{arrows.meta,calc,decorations.pathreplacing}
\usepackage{xspace}
\usepackage{etoc}
\usepackage{algorithm}
\usepackage{algpseudocode}
\setlist{nosep}
\newcommand{\xhdr}[1]{{\vspace{1mm}\noindent{{\bf #1.}}}}

\newcommand{\sysname}{\textsc{Pardon}\xspace}

\title{Ask, Don't Guess: Targeted Clarification for Speech Recognition Errors}
\author{Sankalp Parashar \quad Preethi Jyothi \quad Sunita Sarawagi \\
  IIT Bombay \\
  \texttt{sankalp.parashar.c2024@iitbombay.org}}
\etocdepthtag.toc{mainbody}
\begin{document}
\maketitle

\begin{abstract}
Speech recognition in real time in far-field conversational settings still incurs significant errors. Identifying potential error spans during streaming ASR permits a recourse mechanism where the speaker can be asked to repeat the affected content under cleaner acoustic conditions. We propose \sysname, a new framework for online error localization, that identifies spans likely to benefit from this intervention. Our supervision uses clean audio to estimate how the recognizer's output might change under improved acoustic conditions. We probabilistically align noisy and clean decodes using a pair-HMM to obtain soft supervision. These soft targets train a counterfactual gain scorer by combining a suite of acoustic and language features to produce token-level scores, that are further grouped into spans for targeted recourse. An LLM uses the ASR transcript of the repetition to revise only the selected span. At a precision of $70\%$, \sysname achieves the highest recall by identifying up to 52\% more erroneous tokens compared to the strongest baseline. Targeted repetition yields relative WER reductions of up to 24\% under a budget that accounts for both interruption frequency and the lengths of re-uttered spans.
\end{abstract}

\section{Introduction}

Streaming ASR predicts transcriptions in real time as the model processes incoming speech. However, ambient noise or interfering speakers could obscure parts of the speech signal, leading to localized perception errors. Retrospectively identifying these potential erroneous spans in an evolving transcript is our primary problem of interest. Addressing this problem also has practical value because the online setting naturally supports a recourse mechanism. Once the error spans are localized, a timely request to the speaker to repeat the affected segment under better conditions could improve transcription accuracy.

Error localization has been previously studied in the context of spoken dialogue systems~\citep{skantze2007error} and through confidence estimation modules for streaming ASR~\citep{ragni2018confidence,qiu2021multitask}. Supervision in these prior works typically comes from reference transcripts. However, reference-based supervision targets transcript correctness and not the potential benefit of re-utterance: some errors, such as those arising out of homophones, persist even after re-utterance under clean acoustic condition. This motivates using targets that directly reflect how the recognizer's output might change under cleaner acoustic conditions.

In order to support targeted recourse, we propose a new framework for online error localization in streaming ASR called \sysname, short for \textbf{P}redicting \textbf{A}coustic \textbf{R}e-utterance \textbf{D}esirability \textbf{O}ver \textbf{N}oisy spans. \sysname derives supervision from a \emph{counterfactual estimate} of what the recognizer would output if the same content were heard under cleaner acoustic conditions.
We retrospectively compare these clean decodes with the original noisy decodes by marginalizing over multiple plausible alignments of the clean and noisy decodes
using a pair hidden Markov model (pair-HMM). The pair-HMM assigns each token a soft label given by the posterior probability of a recognition error at that position. These soft labels supervise a counterfactual gain estimator that takes as input both language features and acoustic features extracted via a streaming posterior alignment with audio frames. This estimator runs alongside the ASR's decodes producing token-level scores, which are grouped to identify spans that could benefit from re-utterance.

For recourse, the speaker is asked to repeat the identified span under cleaner acoustic conditions.%
\footnote{We use targeted repetition as an initial recourse strategy. We leave the design of suitable user interfaces, recourse requests beyond repetition and improved policies for user interruptions to future work.}
The ASR model transcribes the repetition and an LLM uses the resulting transcript to revise only the selected span, after which decoding continues.

\noindent
\textbf{Our contributions } can be summarized as follows:
\begin{enumerate*}[(1)]
\item We introduce \sysname, a new framework for online error localization in streaming ASR that isolates potentially erroneous spans through counterfactual supervision from clean audio and a pair-HMM that accounts for uncertainty in the error locations. To the best of our knowledge, the online error localization setting we consider is itself new.
\item We develop a streaming counterfactual gain scorer that combines acoustic and language features, with supervision from the pair-HMM, to identify spans for targeted recourse through repetition.
\item For error localization, we obtain the highest average precision across four challenging
corpora of conversational speech, with the largest margins in the high-precision regime,
where we recall up to $52\%$ more of the erroneous tokens than the strongest baseline at a
precision of 70\%.
\item We harness these localized errors for targeted recourse and reduce the word error rate by up to
$24\%$ relative, and remove up to $30\%$ more of the errors than recent prior work under
the same budget on the number of speaker interruptions and re-uttered words.
\end{enumerate*}

\section{Related Work}
\label{sec:related}
\paragraph{Localizing recognition errors.}
Word-level confidence estimation has a long history in speech
recognition~\citep{jiang2005confidence}. For end-to-end models, confidence is
obtained either directly from the output distribution, where entropy-based
measures detect incorrect words in CTC and RNN-T models considerably better than the
token probability~\citep{laptev2022fast}, or from a learned estimator over the
decoder's internal states~\citep{qiu2021multitask}. Most of this work aims to detect
recognition errors in general. We instead estimate the gain from re-uttering the audio
under cleaner conditions. This shifts our solution in four
ways: how the detector is supervised, which acoustic features it uses, how
evidence is pooled over the transducer's alignment lattice, and how it runs
in a streaming setup alongside standard ASR decoding, where flagged token-level scores are converted to spans for online recourse.

\paragraph{Post-hoc correction.}
Recognition output can also be corrected after the fact, by sequence-to-sequence
correctors~\citep{leng2021fastcorrect} and, more recently, by large language models
that rewrite the transcript from the recognizer's $N$-best
hypotheses~\citep{ma2023can,yang2023generative,chen2023hyporadise}, optionally
conditioned on the noise~\citep{hu2024robustger}. Such correction is complementary to
clarification: it draws on linguistic context rather than new acoustic evidence, and
could be applied on top of a clarified transcript. We include a few-shot prompted $N$-best
corrector as a zero-interruption reference point.

\paragraph{Humans in the loop.}
Prior work on human-assisted ASR correction has largely focused on spoken dialogue systems~\citep{bohus2005sorry,skantze2007error}. These systems serve a single task, such as booking a room or navigation, and treat anything the user
says unrelated to the task as an error. Furthermore, these systems often ask the user to repeat or rephrase the entire turn rather than localizing the error.
More recent work involves users in correcting
open-ended speech: \citet{he2026interactive} let meeting participants correct speaker
attributions with brief feedback, and \citet{peng2026proactive} diagnose the causes of recognition errors and resolve them through multi-turn clarification. Unlike \citet{peng2026proactive}, who seek clarification after each turn, \sysname\ runs
synchronously with the streaming decoder and interrupts the speaker when it identifies a span where repetition is likely to help.

\begin{figure*}[t]
  \centering
  \includegraphics[width=\textwidth]{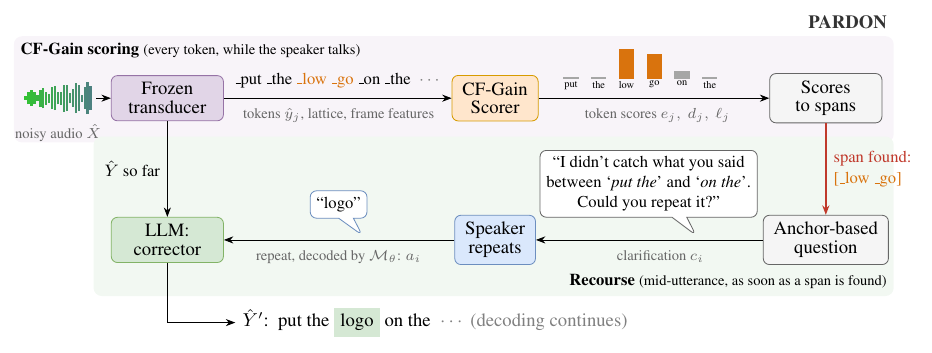}
  \caption{\sysname\ overview. As the speaker talks, the CF-Gain Scorer assigns
  every emitted token a score $e_j$ and two deletion estimates $d_j$ and
  $\ell_j$, and the scores are grouped into spans. Once a span closes,
  \sysname\ asks the speaker to repeat it, using an anchor-based question. An
  LLM corrector puts the repeated words into the transcript.}
  \label{fig:overview}
\end{figure*}

\section{Our Approach}
\label{sec:method}
\vspace{-0.1in}
\newcommand{\vf}{\mathbf{f}}
\newcommand{\real}{\mathbf{R}}
Figure~\ref{fig:overview} shows the pipeline of \sysname. Let $\hat{X}$ denote an audio captured under noisy acoustic conditions, that is
fed to a frozen streaming RNN-transducer $\mathcal{M}_{\theta}$, that produces a
decode $\hat{Y}$ as explained in Section~\ref{sec:rnnt}.
A centerpiece of \sysname\ is a CF-Gain Scorer that after each decoded token $\hat{y}_j$,
synchronously emits a counterfactual gain probability $e_j$ that a
re-utterance of the corresponding audio would improve the transcript, together with two deletion estimates $d_j$ and $\ell_j$
(\S\ref{sec:cfgain}). These scores are used by the recourse module to decide when to ask for a re-utterance; in our initial, simple interface, it pools them into spans as the
transcript grows, and as soon as a noise-corrupted span is identified, asks the speaker to repeat the words in the selected span (\S\ref{sec:spans}). The repetition is decoded by the same transducer
$\mathcal{M}_{\theta}$, and the resulting clarification transcript is passed, together
with the noisy transcript $\hat{Y}$, to an LLM corrector that produces the corrected
transcript $\hat{Y}'$, after which decoding continues (\S\ref{sec:recourse}).

\subsection{Background of RNN-T}
\label{sec:rnnt}
An RNN-T has three components. The encoder maps $\hat{X}$ to states
$h[1], \ldots, h[T]$, one per frame; in a streaming model, $h[t]$ sees no audio
beyond frame $t$ apart from a fixed look-ahead. The predictor encodes the tokens emitted
so far using an RNN : $g[j]$ summarizes $\hat{y}_1, \ldots, \hat{y}_{j-1}$. The joint network combines
the two into a distribution over the vocabulary $\mathcal{V}$ and the blank $\epsilon$:
\begin{equation}
P(\cdot \mid t, j) = \mathrm{softmax}\big(\mathrm{Joint}(h[t], g[j])\big).
\label{eq:joint}
\end{equation}
Greedy decoding scans the frames, emitting the most probable symbol at each step: it
stays at frame $t$ after a token and moves to $t{+}1$ after a blank. We write $t_j$ for
the frame at which $\hat{y}_j$ is emitted.
\subsection{The CF-Gain Scorer}
\label{sec:cfgain}
\begin{figure}[t]
  \centering
  \includegraphics[width=\columnwidth]{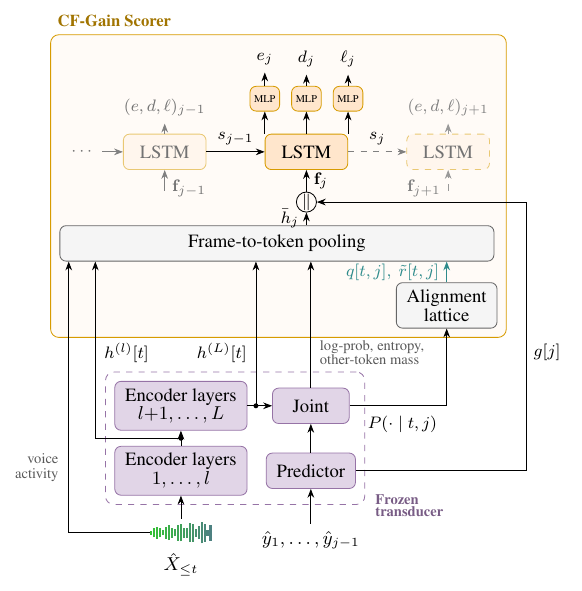}
  \caption{The CF-Gain Scorer (shaded) and the frozen transducer it reads
  (dashed). The emission and dwell posteriors $q$ and $\tilde{r}$ of the
  streaming alignment lattice pool frame-level evidence to each token. The pooled vector $\bar{h}_j$ is
  concatenated with the predictor state $g[j]$ to form $\vf_j$. A unidirectional
  LSTM then scores each token as it is emitted via three heads:
  giving fixable error score $e_j$ and the deletion estimates $d_j$ and $\ell_j$ (\S\ref{sec:probe}).}
  \label{fig:probe}
\end{figure}
After each predicted token from the noisy audio, the CF-Gain scorer estimates the counterfactual probability that a re-utterance of the corresponding audio would lead to a gain in transcription quality.  More precisely, we model $p_\theta(E_j=1|\hat{X},\hat{y}_1,\ldots\hat{y}_j)$ where $E_j$ denotes an error between $j-1$ and $j$ that can be eliminated with a clean re-utterance of the audio after the previous token $\hat{y}_{j-1}$. The error $E_j$ could be either because of missing some token between $\hat{y}_{j-1}$ or $\hat{y}_j$, or because the token $\hat{y}_j$ is wrong. Furthermore, to estimate the amount of error that a re-utterance can recover, we also estimate the number of deletions in the gap between $\hat{y}_{j-1}$ and $\hat{y}_j$, through $d_j$, the probability that an error at $\hat{y}_j$ is a deletion, and $\ell_j$, the number of deleted words in the gap.

To supervise the training of this estimator, we assume a training dataset $\mathcal{D}=\{\hat{X}_i,Y_i,X^c_i:i=1\ldots N\}$ denoting the same reference text $Y_i$ uttered under noisy settings $\hat{X}_i$ and clean settings $X^c_i$.  Such datasets are increasingly available in the public domain, typically obtained by recording conversations with far-field and close-talking microphones, and we experiment with four such in Section~\ref{sec:experiments}.  If clean audio is not available, we show that even synthetically generated TTS audio suffices as distant supervision for our CF-gain scorer.
\begin{figure}[t]
  \centering
  \includegraphics[width=\columnwidth]{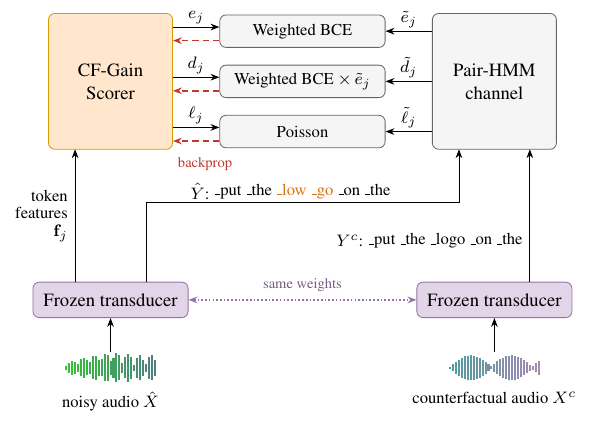}
  \caption{Training supervision. The same frozen transducer decodes the noisy
  audio $\hat{X}$ and the counterfactual audio $X^{c}$. A pair-HMM channel
  aligns the noisy decode $\hat{Y}$ to $Y^{c}$ and gives, for each token, the
  targets $\tilde{e}_j$, $\tilde{d}_j$ and $\tilde{\ell}_j$ of the scorer's three heads;
  only the scorer receives gradients (\S\ref{sec:supervision}).}
  \label{fig:supervision}
\end{figure}
We present the design of the CF-scorer in Section~\ref{sec:features} and describe its training in  Section~\ref{sec:supervision}.

\subsubsection{Design of the CF-Gain scorer}
\label{sec:features}
We implement the CF-Gain scorer $p_\theta(E_j=1|\hat{X},\hat{y}_1,\ldots\hat{y}_j):=e_j$ as another LSTM, much like the LM predictor, with two additional output heads for $d_j$ and $\ell_j$ (Figure~\ref{fig:probe}).
The input to the LSTM is a set of features $\vf_j \in \real^d$ extracted from $\hat{X},\hat{y}_1,\ldots\hat{y}_j$. Instead of processing them afresh, we extract the feature set $\vf_j$ by tapping the current state of the RNN-T.
The features in $\vf_j$ capture evidence from the audio frames, the LM state, and the joint token probabilities.

\xhdr{Features from audio}
At each frame $t$, the encoder reads the audio frames up to $t$ and outputs a hidden
state $h^{(l)}[t]$ at each layer $l$. We extract the
outputs of the final encoder layer and of one intermediate layer.
In addition we extract a voice-activity probability $v[t]$ that frame $t$ contains speech.
We pool these per frame to tokens through the alignment lattice of the RNN-T.
\paragraph{Alignment lattice.}
\label{sec:lattice}
\label{sec:streaming-lattice}
When decoding $\hat{y}_j$ at frame $t_j$, we create a lattice over all possible alignments of the tokens
$\hat{y}_{j'}$, $j' \leq j$, to the frames $t \leq t_j$, by restricting the joint
distribution $P(\cdot \mid t, j')$ (Eq.~\eqref{eq:joint}) to $\hat{y}_{j'}$ and
$\epsilon$. Furthermore, we limit the lattice to a band around the greedy emission path, which limits the alignments to those around the greedy decode (Appendix~\ref{app:lattice}).  Forward--backward over this lattice gives the emission posterior
$q[t, j]$, that $\hat{y}_j$ is emitted at frame $t$, and the dwell posterior
$r[t, j]$, that the decoder advances past frame $t$ while waiting to emit $\hat{y}_j$.

We pool the frame-level audio features to each token under two weightings. The
emission posterior $q[\cdot, j]$ concentrates on the frames at which $\hat{y}_j$ was
emitted. The normalized dwell posterior
$\tilde{r}[t, j] = r[t, j] / \sum_{t'} r[t', j]$ covers the frames the decoder passed
over while waiting to emit $\hat{y}_j$. For layer $l$, the emission view
$\bar{h}^{\mathrm{emit},(l)}_j = \sum_{t} q[t, j]\, h^{(l)}[t]$ describes the audio of
the word that was output, and the dwell view $\bar{h}^{\mathrm{dwell},(l)}_j$, the same
sum weighted by $\tilde{r}$, describes the audio the decoder skipped, which is where a
deleted word would have been. We similarly pool the voice activity score $v[t]$.

\xhdr{Features from the joint network}
We compute a set of scalars from the distribution $P(\cdot \mid t, j)$ output by the
joint network such as its entropy, the log-probability of the emitted token
$\hat{y}_j$, and the other-token mass, the probability of emitting a token
that is neither $\hat{y}_j$ nor the blank. These scalars are defined at every lattice
node $(t, j)$, so we pool them to tokens with $q$ and $\tilde{r}$ as well. Pooling them
in different ways gives 20 scalars, listed in Appendix~\ref{app:features}.

\xhdr{Features from LM}
From the text branch of the transducer, we include $g[j]$ as a feature that summarizes $\hat{y}_1, \ldots, \hat{y}_{j-1}$. This feature is tied to the token, and does not require pooling.

\subsubsection{Training the CF-Gain scorer}
\label{sec:supervision}
 Figure~\ref{fig:supervision} gives an overview of our method of training the gain scorer given dataset  $\mathcal{D}=\{\hat{X}_i,Y_i,X^c_i:i=1\ldots N\}$.
For each instance $i$, we obtain the decode $\hat{Y}_i$ from the noisy audio $\hat{X}_i$, and  the decode $Y^{c}_i = (y^{c}_1, \ldots, y^{c}_I)$ of the counterfactual clean rendering $X^{c}_i$, by the same frozen recognizer
$\mathcal{M}_{\theta}$.  Comparing these sequence-level decodes, we need to get token-level supervision at each $j$.
Since the sequences, can be differing in multiple positions, we marginalize over all possible alignments of the two sequences as follows.

For a fixed  alignment $A$ between $\hat{Y}_i$ and $Y^{c}_i$, each noisy token gets three labels: (1)~
$E_j(A) \in \{0, 1\}$  that denotes if as per this alignment the $j$-th token in
$\hat{Y}_i$ is inserted, substituted, or a token before it is deleted,
(2)~$D_j(A) \in \{0,1\}$ set to 1 if words are deleted before $\hat{y}_j$, and (3)~$L_j(A)$ representing the number of deleted words.
Words deleted after the last token have no following token, so we append an
end-of-sequence position $\hat{y}_{N+1}$ that takes their label.
We marginalize these labels over all possible alignments efficiently using a pair
hidden Markov model~\citep{durbin1998biological}.

A pair-HMM takes as input the two sequences $\hat{Y}_i$ and
$Y^{c}_i$ and gives the posterior probability $P(A \mid Y^{c}_i, \hat{Y}_i)$ of each alignment.
Using an efficient forward--backward algorithm over the pair HMM, we can calculate the marginal probability of the error at $\hat{y}_j$ as
\begin{equation}
\tilde{e}_{ij} = \textstyle\sum_{A} P(A \mid Y^{c}_i, \hat{Y}_i)\, E_j(A).
\end{equation}
We marginalize $D_j(A)$ over the alignments in the same way and divide it by $\tilde{e}_{ij}$, which gives $\tilde{d}_{ij}$, the probability that the error at $\hat{y}_j$ is a deletion.
The pair-HMM requires a set of hyper-parameters, and we fit these via a standard
expectation maximization procedure as described in Appendix~\ref{app:channel}.

\paragraph{Training Objective.}
\label{sec:probe}
To train the scorer, we use weighted binary cross entropy (BCE) losses for the two probabilities and a Poisson loss for the deletion count as follows:
\begin{equation}
\begin{split}
\min_\theta \textstyle\sum_{i,j} \Big[\,&\mathcal{B}_w(\tilde{e}_{ij}, e_{ij}) + \tilde{e}_{ij}\,\mathcal{B}_{w_d}(\tilde{d}_{ij}, d_{ij})\\
&+ \mathbf{1}[\tilde{\ell}_{ij} \geq 1]\,\mathcal{P}(\tilde{\ell}_{ij}, \ell_{ij})\Big]
\end{split}
\label{eq:wbce}
\end{equation}
Here $\mathcal{B}_w(y, p)$ is the weighted BCE loss of the prediction $p$ against the target $y$, and $\mathcal{P}(\tilde{\ell}, \ell)$ is the Poisson loss of $\tilde{\ell} - 1$ with mean $\ell - 1$.
The BCE weights $w$ and $w_d$ are used to upweight the positive term in each BCE loss, respectively, and are estimated such that the balance of positive and negative supervision is equal.
The deletion loss is weighted by $\tilde{e}_{ij}$, so that the deletion head focuses on distinguishing deletions from other error types.

\subsection{Recourse}
\label{sec:recourse}
\subsubsection{From tokens to spans}
\label{sec:spans}
Noise-induced damage tends to extend over several consecutive tokens. It is also more
natural to ask a speaker to repeat a stretch of words than to ask several questions
about separate words. We therefore group the token scores into spans of words as the
transcript is produced. Each token is first assigned a score $m_j$ --- the expected number of errors that a re-utterance would correct at $\hat{y}_j$ and in the gap before it:
\begin{equation}
m_j = e_j\,\big[(1 - d_j) + d_j\,\ell_j\big].
\label{eq:mj}
\end{equation}
An error at $\hat{y}_j$ is either a substitution or insertion with probability $1 - d_j$, that costs one word, or a deletion with probability $d_j$, that costs $\ell_j$ words.
Each word $k$ gets the highest $m_j$ among its tokens, written as $\bar{m}_k$.

We assume that each interruption costs $c_I$ and each re-uttered word costs $c_R$, where each word is re-uttered together with content in the gap before it.
A word that is not selected for repetition costs $\bar{m}_k$, i.e., the expected number of errors left uncorrected at that word and in the preceding gap.  We can use a cumulative sum (CUSUM; \citealp{page1954continuous}) algorithm to compute the best spans in an online setting. More details of the method in Appendix~\ref{sec:app:span}.

\subsubsection{Clarification and correction}
\label{sec:clarification}
The request to repeat a span is a fixed template that quotes the two words on either side of the span, its anchors, and
asks the speaker to restate what they said in between, e.g.\ ``I didn't catch what you
said between `$L$' and `$R$'. Could you repeat it?''. Spans at the start or end of the
utterance have only one anchor and use a shorter template
(Appendix~\ref{app:spans}).

For each span, the LLM corrector receives the transcript decoded up to the
interruption, with the span marked and earlier spans already corrected, together with
the request and the clarification transcript. It returns a replacement for the words of
the span, and the corrected transcript $\hat{Y}'$ leaves
every other word unchanged, so the corrector can only alter the stretches the speaker
was asked about.

\section{Experimental Setup}
\label{sec:setup}
\begin{table*}[t!]
\centering
\small
\setlength{\tabcolsep}{6pt}
\setlength{\aboverulesep}{0pt}
\setlength{\belowrulesep}{0pt}
\renewcommand{\arraystretch}{1.1}
\begin{tabular}{@{}l|ccc|ccc|ccc|ccc@{}}
\toprule
\multirow{3}{*}{Scorer}
& \multicolumn{3}{c|}{AMI}
& \multicolumn{3}{c|}{NOTSOFAR-1}
& \multicolumn{3}{c|}{CHiME-6}
& \multicolumn{3}{c}{DiPCo} \\
\cmidrule(lr){2-4}
\cmidrule(lr){5-7}
\cmidrule(lr){8-10}
\cmidrule(l){11-13}
& \multirow{2}{*}{AP} & \multicolumn{2}{c|}{Recall at prec.}
& \multirow{2}{*}{AP} & \multicolumn{2}{c|}{Recall at prec.}
& \multirow{2}{*}{AP} & \multicolumn{2}{c|}{Recall at prec.}
& \multirow{2}{*}{AP} & \multicolumn{2}{c}{Recall at prec.} \\
\cmidrule(lr){3-4}
\cmidrule(lr){6-7}
\cmidrule(lr){9-10}
\cmidrule(l){12-13}
& & 50 & 70
& & 50 & 70
& & 50 & 70
& & 50 & 70 \\
\midrule
Entropy
& 38.7 & 23.8 & 4.0
& 39.2 & 28.6 & 6.3
& 51.0 & 49.7 & 18.5
& 41.3 & 29.0 & \textbf{11.0} \\
Qiu et al.
& 55.1 & 58.5 & 25.3
& 56.5 & 61.6 & 25.9
& 65.7 & 80.0 & 44.4
& 42.1 & 33.0 & 2.5 \\
Wang et al.
& 52.9 & 56.3 & 18.3
& 50.5 & 51.1 & 13.7
& 62.4 & 75.5 & 36.1
& 35.4 & 9.2 & 0.0 \\
TruCLeS
& 47.7 & 41.9 & 13.1
& 45.5 & 35.2 & 8.8
& 58.6 & 67.2 & 30.0
& 32.3 & 1.9 & -- \\
Peng et al.
& 47.2 & 46.0 & 3.8
& 49.5 & 54.5 & 7.8
& 57.1 & 72.6 & 11.2
& 41.1 & 25.9 & 1.7 \\
\midrule
CF-Gain
& \textbf{56.5} & \textbf{61.7} & \textbf{32.5}
& \textbf{59.8} & \textbf{68.7} & \textbf{39.4}
& \textbf{66.7} & \textbf{80.3} & \textbf{49.4}
& \textbf{44.1} & \textbf{36.6} & 10.0 \\
\bottomrule
\end{tabular}
\caption{Token-level scoring on the eval set of each corpus,
against labels from the close-talking decode. Trained scorers are
trained on the same corpus; the CF-Gain Scorer is evaluated by $e_j$. Share of erroneous tokens (the AP of a
random scorer): AMI 21.5, NOTSOFAR-1 16.4, CHiME-6 24.8,
DiPCo 19.3. Best result for each metric within each corpus is
in bold; -- indicates that the target precision was never reached.}
\label{tab:probe-accuracy}
\end{table*}

\paragraph{ASR System.}
All experiments use Nemotron Speech Streaming~\citep{nvidia2026nemotron}, a
0.6B-parameter English transducer with a FastConformer~\citep{rekesh2023fast} encoder of 24 layers,
a 1024-dimensional encoder output at a frame rate of 80\,ms, and a vocabulary of 1024
subword tokens plus the blank. We use its default attention context of 70 frames to the
left and 13 to the right, so the encoder consumes audio in chunks of 1.12\,s with at
most 1.04\,s of look-ahead. The transducer is frozen throughout.

\paragraph{Datasets.}
We use four corpora of conversational speech: two of them contain meeting conversations,
AMI~\citep{carletta2005ami} and NOTSOFAR-1~\citep{vinnikov2024notsofar}, and two of them
contain dinner party conversations, CHiME-6~\citep{watanabe2020chime6} and
DiPCo~\citep{vansegbroeck2020dipco}. All four corpora contain both far-field channels and a close-talking recording of every speaker (a headset in AMI, a close-talking microphone in NOTSOFAR-1
and DiPCo, and a binaural microphone in CHiME-6). We use the official session-level train/development/evaluation splits~\citep{amipartitions,cornell2024chime8}. Each utterance is a speaker turn extracted from both far-field and close-talking recordings using reference timestamps. Very short turns and turns containing reference words marked as unintelligible are excluded from training; development and evaluation sets retain all turns. The development (dev) set is
used for model selection and calibration, and all numbers are reported on the
evaluation (eval) set.

To show how \sysname performs in the absence of a parallel close-talking recording, we also create a synthetic rendering of each training utterance by synthesizing its reference transcript with
a text-to-speech (TTS) model, Kokoro-82M~\citep{kokoro2025}.
Dev and eval labels come from
the close-talking recordings. Appendix~\ref{app:data} describes the corpora, how the turns
are cut, speech synthesis details and the train/dev/eval set sizes after segmentation.

\subsection{Implementation Details}
\label{sec:implementation}

We derive supervision by aligning normalized noisy and clean decodes by the same ASR system using a pair-HMM, and copying word-level labels to its subwords. Token features combine emission- and dwell-pooled encoder states from 13th and 24th blocks with the predictor state and scalar features of \S\ref{sec:features}; voice activity
probabilities come from Silero VAD~\citep{silero2021vad}. The CF-Gain Scorer uses a single-layer unidirectional LSTM with 128 hidden units and 3 prediction heads of \S\ref{sec:probe}. Remaining implementation details are in Appendix~\ref{app:implementation}.

\subsection{Baselines}
\label{sec:baselines}

\paragraph{Token-level scoring.}
We compare the CF-Gain Scorer with five baseline scorers: an entropy-based method and four learned confidence estimators.
\begin{enumerate*}[label=\roman*)]
    \item \citet{laptev2022fast}: Tsallis entropy of the recognizer's output distribution at token emission, with $\alpha$ tuned on the development set.
    \item \citet{qiu2021multitask}: a Transformer over hypothesis tokens that attends
to encoder frames and predicts token errors and deletions.
    \item \citet{wang2021rnnt}: a Transformer over token embeddings and encoder frames at emission.
    \item TruCLeS~\citep{ravi2025trucles}: an LSTM predicting  confidence based on lexical similarity to the reference.
    \item \citet{peng2026proactive}: detects errors from the
recognizer's internal states with separate perception-error and deletion detectors, whose scores we combine. We omit their comprehension-error detector from our experiments as it targets errors not repairable by clean re-utterance.
\end{enumerate*}
Every estimator is trained on re-utterance gains from close-talking decodes rather than
reference-based labels, on the same data as the CF-Gain Scorer, and adapted to streaming, using only input available at token emission
(Appendix~\ref{app:token-baselines}).

\paragraph{Recourse.} We compare \sysname with two baselines: one without user clarification and the other using it exactly the same way as \sysname.
\begin{enumerate*}[label=\roman*)]
\item $N$-best correction: Following HyPoradise~\citep{chen2023hyporadise}, an LLM corrects the transcript from a
5-best list: the greedy transcript and the four best distinct hypotheses from a width-32 beam
search. We use its hypotheses-to-transcription (H2T)
prompt with Gemini 3.8 Flash and 10 examples from the dev set
(Appendix~\ref{app:prompt-nbest}). We also report the oracle that picks the best
hypothesis from the list.
\item Proactive ASR~\citep{peng2026proactive}: We use the same recourse pipeline as our system, with \citet{peng2026proactive} used for localizing the errors.
\end{enumerate*}

\subsection{Evaluation metrics}
\label{sec:evaluation}
 \begin{table}[t]
\centering
\small
\setlength{\tabcolsep}{3.5pt}
\resizebox{\columnwidth}{!}{%
\begin{tabular}{lllccc}
\toprule
 & & & & \multicolumn{2}{c}{Recall at prec.} \\
\cmidrule(lr){5-6}
Test & Scorer & Trained on & AP & 50 & 70 \\
\midrule
\multirow{3}{*}{NOTSOFAR-1} & Entropy & -- & 39.2 & 28.6 & 6.3 \\
                            & CF-Gain & NOTSOFAR-1 & 59.8 & 68.7 & 39.4 \\
                            & CF-Gain & AMI        & \textbf{63.2} & \textbf{71.7} & \textbf{39.7} \\
\midrule
\multirow{3}{*}{DiPCo}      & Entropy & -- & 41.3 & 29.0 & 11.0 \\
                            & CF-Gain & DiPCo   & 44.1 & 36.6 & 10.0 \\
                            & CF-Gain & CHiME-6 & \textbf{54.8} & \textbf{58.0} & \textbf{24.4} \\
\bottomrule
\end{tabular}}
\caption{A CF-Gain Scorer trained on one corpus is applied
unchanged to the other corpus of the same genre, and compared with a scorer trained on
the target corpus along with the training-free entropy baseline.
Best result per block in bold.}
\label{tab:ood}
\end{table}

\paragraph{Token-level scoring.}
Token scores $e_j$ are evaluated on the eval set of each corpus against the hard
labels $E_j$ from the close-talking decode, using average precision (AP) and
recall at precisions of 0.5 and 0.7. We also report the share of erroneous tokens, which is the AP of a random scorer.

\paragraph{Recourse.}

Recourse is evaluated by the relative WER reduction over the far-field greedy transcript against gold references, under a \emph{re-utterance burden} $B = n_I + \beta\, n_W$. Here, $n_I$ and $n_W$ count interruptions and re-uttered words per minute of audio, respectively; we use $\beta \in \{0.05, 0.10\}$. The speaker's reply is simulated by the close-talking recording of the span or by TTS of
its reference words. Further recourse details are in Appendix~\ref{app:spans}, some recourse examples are in Appendix~\ref{app:rec-examples}, and tuning details for
costs $(c_R, c_I)$ on dev set are in Appendix~\ref{app:costtuning}.

 \section{Results and Analysis}
\label{sec:results}
\label{sec:experiments}

\subsection{Token-level scoring}
\label{sec:loc-results}

\paragraph{Scoring accuracy.}
Table~\ref{tab:probe-accuracy} compares the CF-Gain Scorer with the five scoring
baselines on each corpus. The CF-Gain Scorer has the highest AP on all four corpora scoring
\emph{1.0 to 3.3 points} above the strongest baseline, \citet{qiu2021multitask}. The margin is
\emph{larger at high precision}: at a precision of 70\%, it recalls 7.2, 13.5 and 5.0 points more of the erroneous tokens than the best baseline on AMI, NOTSOFAR-1 and CHiME-6.

\paragraph{Out-of-domain generalization.}
To check how well the CF-Gain Scorer performs out of domain, we apply a scorer trained
on one corpus unchanged to the other corpus of the same type: from AMI to NOTSOFAR-1 for
meetings, and from CHiME-6 to DiPCo for dinner parties (Table~\ref{tab:ood}). On
NOTSOFAR-1, the scorer trained on AMI exceeds the in-domain scorer on every metric by
3.4 AP points. On DiPCo, the
scorer trained on CHiME-6 exceeds the in-domain scorer on every metric by 10.7 AP points. In both cases, the source corpus has far more training data than the target (62.9
vs.\ 6.2 hours and 28.5 vs.\ 0.7 hours; Table~\ref{tab:data-stats}), which suggests that
a scorer trained on a larger corpus of the same genre can replace one trained in-domain. With a subset of the source as large as the target's training set, the transferred
scorer still stays within 4 AP points of the in-domain one (Appendix~\ref{app:ood-subset}).
\begin{table}[t]
\centering
\small
\setlength{\tabcolsep}{4pt}
\begin{tabular}{llccc}
\toprule
 & & & \multicolumn{2}{c}{Recall at prec.} \\
\cmidrule(lr){4-5}
Corpus & Supervision & AP & 50 & 70 \\
\midrule
\multirow{2}{*}{AMI}        & Close-talk & \textbf{56.5} & \textbf{61.7} & \textbf{32.5} \\
                            & TTS        & 54.7 & 59.1 & 27.4 \\
\midrule
\multirow{2}{*}{NOTSOFAR-1} & Close-talk & \textbf{59.8} & \textbf{68.7} & \textbf{39.4} \\
                            & TTS        & 51.5 & 54.8 & 15.9 \\
\midrule
\multirow{2}{*}{CHiME-6}    & Close-talk & \textbf{66.7} & 80.3 & \textbf{49.4} \\
                            & TTS        & 65.3 & \textbf{81.0} & 44.7 \\
\midrule
\multirow{2}{*}{DiPCo}      & Close-talk & \textbf{44.1} & \textbf{36.6} & \textbf{10.0} \\
                            & TTS        & 43.7 & 34.1 & 8.2 \\
\bottomrule
\end{tabular}
\caption{CF-Gain Scorers trained with labels from the close-talking recording or from a
synthesized (TTS) rendering of the reference, both evaluated against close-talk labels
on the evaluation set.
Best result per corpus in bold.}
\label{tab:tts}
\end{table}
\paragraph{Synthetic supervision.}
We also train a scorer on each corpus with labels from TTS of the
reference (Appendix~\ref{app:tts}) instead of the close-talking recording, and evaluate it against the close-talk labels (Table~\ref{tab:tts}). On AMI, CHiME-6 and DiPCo, TTS
loses at most 1.8 points of AP and 5.1 points of recall at 70\% precision. On AMI and CHiME-6 it still recalls more at 70\%
precision than every baseline of Table~\ref{tab:probe-accuracy}, all  trained with close-talk labels, and on DiPCo it still has a higher AP than all of them.
NOTSOFAR-1 is the exception: there it loses 8.3 points of AP and 23.5 points of
recall at 70\% precision, and falls below \citet{qiu2021multitask}.

\paragraph{Ablations.}
Table~\ref{tab:ablation} removes one group of features at a time from the CF-Gain
Scorer, on AMI and CHiME-6. The intermediate encoder layer and the joint-network scalars
both matter. Without layer 13, recall at 70\% precision drops by 8.4 points on AMI and
by 3.1 points on CHiME-6. Without the joint-network scalars, it drops by 4.3 and 8.4
points. Limiting frame level features to the emission frame based on greedy decoding has a smaller effect: AP drops by 0.8 points on AMI and 0.2 points on CHiME-6.%

\subsection{Recourse}
\label{sec:recourse-results}

  \begin{figure}[!t]
  \centering
  \includegraphics[width=0.9\columnwidth]{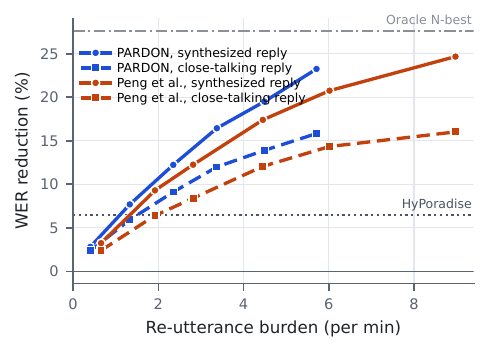}\\[1pt]
  {\small\textbf{(a) AMI}}\\[4pt]
  \includegraphics[width=0.9\columnwidth]{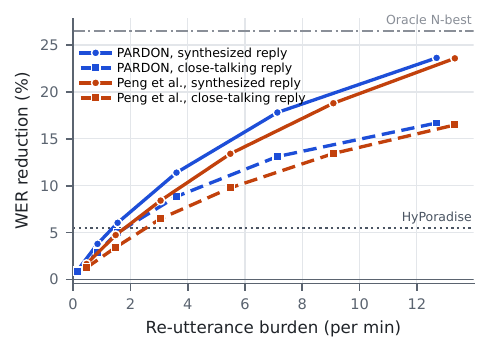}\\[1pt]
  {\small\textbf{(b) CHiME-6}}
  \caption{Recourse on the AMI (a) and CHiME-6 (b) eval sets ($\beta = 0.05$, spans of
  at most six words): relative WER reduction against the re-utterance burden with synthesized (solid) or
  close-talking (dashed) replies. \sysname\ uses the CF-Gain Scorer; Peng et al.\ uses the
  same recourse pipeline with the detectors of \citet{peng2026proactive}.
  Horizontal lines: the zero-interruption $N$-best correction of HyPoradise and the $N$-best oracle. We also report WER reductions achieved through speech
  enhancement (which is quite poor) in Appendix~\ref{app:enhancement}.}
  \label{fig:recourse}
\end{figure}
Figure~\ref{fig:recourse} shows the WER reduction of
\sysname\ and of the same pipeline driven by \citet{peng2026proactive} against the
re-utterance burden, for spans of at most six words and $\beta = 0.05$ on AMI and CHiME-6.
On AMI, \sysname\ removes 23\% of the word errors with synthesized replies at a
burden of 6 per minute, where Peng et al.\ needs 8, and 16\% with close-talking replies
at 6 per minute, where Peng et al.\ needs 9. On CHiME-6, whose noisy decode has about
twice the WER of AMI (64\% vs.\ 30\%), the same reductions take higher burdens: \sysname\
removes 18\% and 13\% at 7 per minute, where Peng et al.\ needs 8 and 9.

Exceeding the zero-interruption $N$-best correction of HyPoradise (6\% on AMI and 5\% on
CHiME-6) requires a burden of around 2 per minute on both corpora, with either kind of
reply. At the highest burdens, \sysname\ with synthesized replies comes within 4 points
of the $N$-best oracle (28\% on AMI and 27\% on CHiME-6), which no zero-interruption
method can exceed with these hypotheses. Synthesized replies remove more errors than
close-talking ones, by up to 7 points, likely because
the close-talking recording of a span
also carries disfluencies and crosstalk from the other speakers and is decoded with more
errors. The ordering of the two scorers is the same with $\beta = 0.10$ and with spans of
up to eight words (Appendix~\ref{app:recourse-more}).

\section{Conclusion}
\label{sec:conclusion}

 \begin{table}[t]
\centering
\small
\setlength{\tabcolsep}{3.5pt}
\setlength{\aboverulesep}{0pt}
\setlength{\belowrulesep}{0pt}
\renewcommand{\arraystretch}{1.05}
\resizebox{\columnwidth}{!}{%
\begin{tabular}{@{}l|ccc|ccc@{}}
\toprule
\multirow{3}{*}{Features}
& \multicolumn{3}{c|}{AMI}
& \multicolumn{3}{c}{CHiME-6} \\
\cmidrule(lr){2-4}
\cmidrule(l){5-7}
& \multirow{2}{*}{AP} & \multicolumn{2}{c|}{Recall at prec.}
& \multirow{2}{*}{AP} & \multicolumn{2}{c}{Recall at prec.} \\
\cmidrule(lr){3-4}
\cmidrule(l){6-7}
& & 50 & 70 & & 50 & 70 \\
\midrule
CF-Gain (all features)& \textbf{56.5} & \textbf{61.7} & 32.5 & \textbf{66.7} & 80.3 & \textbf{49.4} \\
\midrule
Emission frame only& 55.7 & 60.5 & \textbf{33.4} & 66.5 & \textbf{81.4} & 46.9 \\
w/o layer 13& 54.2 & 58.7 & 24.1 & 65.7 & 78.4 & 46.3 \\
w/o joint scalars& 54.8 & 59.5 & 28.2 & 63.7 & 77.6 & 41.0 \\
\bottomrule
\end{tabular}}
\caption{Feature ablations of the CF-Gain Scorer on the evaluation sets of AMI and
CHiME-6. \emph{Emission frame only}: uses encoder features only from the greedy emission frame.
\emph{w/o layer 13}: only the final encoder layer is used.
\emph{w/o joint scalars}: the scalar features derived from the joint network are
removed. All numbers are in percent; best per column in bold.}
\label{tab:ablation}
\end{table}
We introduced \sysname, which improves streaming recognition under noise by asking the
speaker to repeat the stretches it has likely misheard. Its CF-Gain Scorer reads its
evidence from the transducer's own alignment lattice, is trained on soft labels that
marginalize over the alignment between the noisy and the counterfactual clean decode,
and estimates how many words a repetition would recover. On four corpora, it ranks
erroneous tokens above entropy-based and learned confidence estimators, and
it transfers unchanged to another corpus of the same genre. With these scores, \sysname\
removes 23\% of the word errors on far-field AMI meetings with synthesized replies, and
16\% with close-talking replies, at a re-utterance burden of under 6 per minute, more
than both an $N$-best LLM corrector and the same pipeline driven by
\citet{peng2026proactive}. Evaluating clarification with real speakers instead of
simulated replies is the natural next step.

\section*{Limitations}
We study a single recourse mechanism, targeted repetition. In realistic interactions,
users may respond in many other ways, for instance by paraphrasing, correcting only part
of the span or adding context, which would make the LLM corrector's task harder, and our
replies are simulated rather than collected from real speakers. The scoring step is
designed for one family of streaming recognizers, RNN transducers, whose alignment
lattice it reads; other architectures would need a different source of token-level
evidence. Finally, we focus on errors caused by far-field recording, because the
parallel close-talking audio our supervision needs is readily available in such corpora;
the same counterfactual supervision could be applied to other kinds of ASR errors
wherever a cleaner rendering of the same speech can be obtained.

\section*{Ethical Considerations}
All four corpora (AMI, NOTSOFAR-1, CHiME-6 and DiPCo) are publicly released for research
and were recorded with the consent of their participants; we use them under their
licenses and do not redistribute any audio. No new data was collected from people:
speaker replies are simulated with the existing close-talking recordings or with
synthesized speech.

A deployed system raises two risks. First, recognition errors are more frequent for
speakers whose accents or speaking styles are under-represented in training data, so
such speakers may be interrupted more often and bear more of the burden of correction;
we have not measured this, and it should be evaluated before deployment. Second, the LLM
corrector can insert words the speaker did not say, which matters when transcripts serve
as records, such as meeting minutes; a deployed system should mark corrected spans.
\bibliography{references}

\newpage
\appendix
\etocdepthtag.toc{appendix}

\begingroup
  \etocsettagdepth{mainbody}{none}
  \etocsettagdepth{appendix}{subsection}
  \etocsettocstyle{\section*{Contents of the Appendix}}{}
  \tableofcontents
\endgroup

\section{Transducer Alignment Lattice: Definitions}
\label{app:lattice}
For an utterance $\hat{X}$ of $T$ frames $(\hat{x}_1, \ldots, \hat{x}_T)$ and a
transcript $\hat{Y}$ of $N$ tokens $(\hat{y}_1, \ldots, \hat{y}_N)$, an alignment is a
monotonic path through a $T \times (N{+}1)$ grid whose node $(t, j)$ is the state in
which the first $t$ frames $\hat{x}_1, \ldots, \hat{x}_t$ have been processed,
$\hat{y}_1, \ldots, \hat{y}_{j-1}$ have been emitted, and the decoder waits to emit
$\hat{y}_j$; row $N{+}1$ follows the last token. We consider only the paths that run
from $(1, 1)$ to $(T, N{+}1)$. They differ only in \emph{when} each token is emitted,
and the probability of the transcript is the total probability of all of them,
\begin{equation}
P(\hat{Y} \mid \hat{X}) \;=\; \sum_{\text{paths } (1,1) \to (T, N+1)}
P(\text{path} \mid \hat{X}).
\end{equation}
At each node the model produces a distribution over $\mathcal{V} \cup \{\epsilon\}$,
where $\mathcal{V}$ is the vocabulary of the model and $\epsilon$ the blank,
\begin{equation}
P(\cdot \mid t, j) = \mathrm{softmax}\big(\mathrm{Joint}(h[t], g[j])\big),
\end{equation}
where $h[t]$ is the encoder state at frame $t$, conditioned on the audio up to that
frame, and $g[j]$ the predictor state, conditioned on $\hat{y}_1, \ldots, \hat{y}_{j-1}$.

Everything that follows is computed on a lattice \emph{restricted} to the 1-best
transcript $\hat{Y}$ obtained by greedy decoding. Although the model assigns
probability to every element of $\mathcal{V}$ at every node, the restricted lattice
admits only the token that continues $\hat{Y}$. Only two transitions therefore leave
$(t, j)$: emitting $\epsilon$, which advances the frame, $(t, j) \rightarrow (t{+}1, j)$,
and emitting $\hat{y}_j$, which advances the token index, $(t, j) \rightarrow (t, j{+}1)$,
with probabilities
\begin{equation}
b[t, j] = P(\epsilon \mid t, j),
\quad
e[t, j] = P(\hat{y}_j \mid t, j).
\end{equation}

The forward variable $\alpha[t, j]$ is the probability of reaching $(t, j)$ and the
backward variable $\beta[t, j]$ the probability of completing the alignment from it:
\begin{align}
\alpha[1, 1] &= 1, \\
\alpha[t, j] &= \alpha[t{-}1, j]\,b[t{-}1, j] \nonumber \\
             &\quad + \alpha[t, j{-}1]\,e[t, j{-}1], \\
\beta[T, N{+}1] &= b[T, N{+}1], \\
\beta[t, j]  &= \beta[t{+}1, j]\,b[t, j] \nonumber \\
             &\quad + \beta[t, j{+}1]\,e[t, j],
\end{align}
with out-of-lattice terms taken as zero, giving
$Z = P(\hat{Y} \mid \hat{X}) = \alpha[T, N{+}1]\,b[T, N{+}1] = \beta[1, 1]$.

\paragraph{Emission probability.}
$q[t, j]$ is the posterior probability that token $\hat{y}_j$ is emitted at frame
$t$, given that $\hat{Y}$ is the final transcript; it is the occupancy of the label
transition out of $(t, j)$:
\begin{equation}
q[t, j] \;=\; \frac{\alpha[t, j]\;e[t, j]\;\beta[t, j{+}1]}{Z}.
\label{eq:q}
\end{equation}

\paragraph{Dwell probability.}
$r[t, j]$ is the posterior probability, given that $\hat{Y}$ is the final transcript,
that the decoder advances from frame $t$ to frame $t{+}1$ while waiting to emit
$\hat{y}_j$, i.e.\ in the gap between $\hat{y}_{j-1}$ and $\hat{y}_j$ (the frame at
which $\hat{y}_{j-1}$ is emitted included); it is the occupancy of the blank transition
out of $(t, j)$:
\begin{equation}
r[t, j] \;=\; \frac{\alpha[t, j]\;b[t, j]\;\beta[t{+}1, j]}{Z}.
\label{eq:r}
\end{equation}

\paragraph{Normalization.}
The two quantities normalize along different axes. Each token is emitted exactly once
in any alignment, so $\sum_{t} q[t, j] = 1$ and $q[\cdot, j]$ is a distribution over
frames. In contrast, because each frame is advanced exactly once,
$\sum_{j} r[t, j] = 1$: $r[\cdot, j]$ instead sums to the expected number of blank
frames between $\hat{y}_{j-1}$ and $\hat{y}_{j}$, which may exceed one or approach
zero. When pooling frame-level features to token level we therefore divide
$r[\cdot, j]$ by its sum over frames.

\paragraph{Streaming estimation.}
The posteriors above use the whole utterance. In \sysname\ the posteriors of token
$\hat{y}_j$ are computed when the greedy decoder emits it, at frame $t_j$. Let $N_j$ be
the number of tokens the decoder has emitted by frame $t_j$ (at least $j$, more if
further tokens are emitted at the same frame). The truncated backward variable
$\beta_j$ follows the recursion above on frames $1, \ldots, t_j$ and rows
$1, \ldots, N_j{+}1$, with the terminal condition
\begin{align}
\beta_j[t_j{+}1, N_j{+}1] &= 1, \\
\beta_j[t_j{+}1, i] &= 0 \quad (i \leq N_j),
\end{align}
so that every path has emitted the $N_j$ known tokens by the cut, and
$Z_j = \beta_j[1, 1]$. The forward variable needs no change, since it depends only on
earlier frames. The posteriors of token $j$ are then
\begin{align}
q_j[t, j] &= \alpha[t, j]\,e[t, j]\,\beta_j[t, j{+}1] \,/\, Z_j, \\
r_j[t, j] &= \alpha[t, j]\,b[t, j]\,\beta_j[t{+}1, j] \,/\, Z_j,
\end{align}
for $t \leq t_j$, and zero for later frames. When $t_j = T$ this is the full lattice.
The end-of-sequence position $\hat{y}_{N+1}$ is never emitted, so it has no emission
frame; its posteriors are computed once the utterance has ended, on the full lattice,
from row $N{+}1$ (the frames after the last token).

\paragraph{Band.}
Both recursions are further restricted to the nodes $(t, j)$ with
\begin{equation}
t_{j-1} - \delta \;\leq\; t \;\leq\; t_{j+k},
\end{equation}
where the lower bound is $1$ for $j = 1$ and the upper bound is $T$ when $j + k > N$;
$\alpha$ and $\beta_j$ are zero outside the band. Since node $(t, j)$ is reached only
once $\hat{y}_{j-1}$ has been emitted, a path may thus emit each token at most $\delta$ frames before the decoder did, and may fall at most $k$ tokens behind it. We use $k = 3$
and $\delta = 16$ frames.
\section{Token Features}
\label{app:features}
The input $\vf_j$ of the CF-Gain Scorer concatenates four pooled encoder vectors (the
emission and dwell views of \S\ref{sec:features} at the intermediate and the final
encoder layer), the predictor state $g[j]$, and the 27 scalars below: 20 computed from
the joint network's output distribution (1--20) and 7 from the voice-activity
probability $v[t]$ (21--27). Every feature is computed from the streaming posteriors of
\S\ref{sec:streaming-lattice}, so it uses only frames up to the emission of
$\hat{y}_j$; features are standardized with training-set statistics.

\paragraph{Notation.}
$P_t = P(\cdot \mid t, j)$ is the output distribution over
$\mathcal{V} \cup \{\epsilon\}$ at node $(t, j)$, and $e[t, j]$ and $b[t, j]$ are the
probabilities of the emitted token and of the blank (Appendix~\ref{app:lattice}). Let
$o[t, j] = 1 - e[t, j] - b[t, j]$ be the probability of any other token. The emission
frames of $\hat{y}_j$ are $\mathcal{T}^{\text{emit}}_j = \{t : q[t, j] > 0.01\}$, and its dwell frames are
$\mathcal{T}^{\text{gap}}_j = \{t : r[t, j] \geq 0.01 \max_{t'} r[t', j]\}$, the frames of the gap before it;
$\tilde{r}[t, j]$ is the normalized dwell posterior. The voice-activity probability
$v[t]$ is the unweighted mean of the Silero VAD~\citep{silero2021vad} outputs of the
three 32\,ms chunks that overlap the 80\,ms encoder frame, including a chunk that
overlaps it only partly. For a distribution $p$,
$H(p) = -\sum_k p_k \log p_k$ is its Shannon entropy and
$S_\alpha(p) = (1 - \sum_k p_k^{\alpha}) / (\alpha - 1)$ its Tsallis entropy. A statistic
over an empty frame set is $0$.

\paragraph{Uncertainty at emission (5).}
How spread out the model's output is when it emits $\hat{y}_j$, averaged over the
frames at which it may have been emitted:
\begin{list}{}{\setlength{\leftmargin}{3.3em}\setlength{\labelwidth}{3em}\setlength{\labelsep}{0.3em}\setlength{\itemsep}{1pt}\setlength{\parsep}{0pt}\setlength{\topsep}{2pt}}
  \item[(1)] $\sum_t q[t, j]\, H(P_t)$
  \item[(2--4)] $\sum_t q[t, j]\, S_\alpha(P_t)$ for $\alpha = 0.5, 2, 4$
  \item[(5)] $\sum_t q[t, j] \log e[t, j]$
\end{list}
The Shannon entropy weighs all alternatives; the Tsallis entropies, which
\citet{laptev2022fast} found well suited to transducer confidence, shift the weight
from the many low-probability alternatives ($\alpha < 1$) to the few strongest
competitors ($\alpha > 1$). The expected log-probability (5) is the usual token
confidence.

\paragraph{Probability of other tokens (3).}
The probability the model gives to tokens other than $\hat{y}_j$ and the blank,
averaged under three weightings:
\begin{list}{}{\setlength{\leftmargin}{3.3em}\setlength{\labelwidth}{3em}\setlength{\labelsep}{0.3em}\setlength{\itemsep}{1pt}\setlength{\parsep}{0pt}\setlength{\topsep}{2pt}}
  \item[(6)] $\sum_t q[t, j]\, o[t, j]$, at the emission
  \item[(7)] $\sum_t \tilde{r}[t, j]\, o[t, j]$, over the gap
  \item[(8)] $\sum_t w[t, j]\, o[t, j] \,/\, \sum_t w[t, j]$, over both, with
  $w = q[\cdot, j] + r[\cdot, j]$
\end{list}
At the emission, mass on other tokens indicates that a different word may have been
spoken (a substitution). In the gap, where the decoder emitted only blanks, it
indicates that the model considered emitting a word there and did not, the trace a
deleted word leaves.

\paragraph{Frame statistics (12).}
The weighted averages above can hide a single frame at which the model strongly
preferred something else. We therefore also take, without weighting, the mean, maximum
and 90th percentile of two per-frame values over each frame set:
\begin{list}{}{\setlength{\leftmargin}{3.3em}\setlength{\labelwidth}{3em}\setlength{\labelsep}{0.3em}\setlength{\itemsep}{1pt}\setlength{\parsep}{0pt}\setlength{\topsep}{2pt}}
  \item[(9--11)] $\log e[t, j]$ over $t \in \mathcal{T}^{\text{emit}}_j$
  \item[(12--14)] $o[t, j]$ over $t \in \mathcal{T}^{\text{emit}}_j$
  \item[(15--17)] $\log e[t, j]$ over $t \in \mathcal{T}^{\text{gap}}_j$
  \item[(18--20)] $o[t, j]$ over $t \in \mathcal{T}^{\text{gap}}_j$
\end{list}
Over the emission frames, they describe how confidently and how consistently
$\hat{y}_j$ was preferred. Over the gap, (15--17) show whether $\hat{y}_j$ was already
likely before it was emitted, and (18--20) whether some other token was, at any point
in the gap.

\paragraph{Voice activity (7).}
Whether the frames contain speech at all, independently of the recognizer:
\begin{list}{}{\setlength{\leftmargin}{3.3em}\setlength{\labelwidth}{3em}\setlength{\labelsep}{0.3em}\setlength{\itemsep}{1pt}\setlength{\parsep}{0pt}\setlength{\topsep}{2pt}}
  \item[(21)] $\sum_t \tilde{r}[t, j]\, v[t]$, over the gap
  \item[(22--24)] mean, maximum and 90th percentile of $v[t]$ over $t \in \mathcal{T}^{\text{emit}}_j$
  \item[(25--27)] mean, maximum and 90th percentile of $v[t]$ over $t \in \mathcal{T}^{\text{gap}}_j$
\end{list}
Speech in a gap through which the decoder emitted only blanks suggests that words
were spoken there and lost; little speech around an emitted token suggests that it was
produced from noise rather than from a word.

\smallskip\noindent
The end-of-sequence position $\hat{y}_{N+1}$ is treated as a token emitted at the last
frame: its emission frame set is that frame, its dwell frames are those after
$\hat{y}_N$, and since only the blank can be emitted from row $N{+}1$, the probability
of the final blank takes the place of $e$, with $e[t, N{+}1] = b[t, N{+}1]$ and
$o[t, N{+}1] = 1 - b[t, N{+}1]$.
\section{\sysname\ Algorithm}
\label{app:algorithm}
Algorithm~\ref{alg:pardon} summarizes the training of the CF-Gain Scorer (\S\ref{sec:cfgain})
and the streaming inference with span selection and recourse (\S\ref{sec:recourse}).

\subsection{Online spans with the CUSUM algorithm}
\label{sec:app:span}
Re-uttering word $k$ yields a net gain of $\bar{m}_k - c_R$, and a span of words warrants
an interruption when these gains sum to at least $c_I$. We track the gains with a
cumulative sum (CUSUM; \citealp{page1954continuous}):
\begin{equation}
C_k = \max\big(0,\; C_{k-1} + \bar{m}_k - c_R\big), \quad C_0 = 0.
\label{eq:cusum}
\end{equation}
$C_k$ is the largest total gain of a span of words that ends at word $k$ and starts
after the last interruption.
Whenever the cumulative gain becomes positive, we track its current value $C_k$, its
maximum $P$ during the positive run and the word where $P$
was reached. Once $P - C_k \geq c_I$, we interrupt and select the span from the first word of the positive run to where $P$ was
reached. Words after the peak are not part of the span; they only ensure that we do not
interrupt while the speech remains degraded. After the interruption, the cumulative gain resets to zero. If $C_k$ returns to zero before $P - C_k$ reaches $c_I$, no span is formed. If
the turn ends first, we interrupt if $P \geq c_I$.

\begin{algorithm*}[t]
\caption{\sysname\ end to end. \textsc{Train} (offline) fits the CF-Gain Scorer $p_\theta$
on a dataset $\mathcal{D}$ of noisy and clean renderings of the same utterances; \textsc{Infer}
runs alongside the streaming decode of $\mathcal{M}_{\theta}$. The pair HMM has parameters
$\Phi = (\pi, \gamma, \eta, \rho)$:
the probability that a noisy word is an error ($\pi$), that a run of deleted words
continues ($\gamma$), that the error after a deletion is its anchor ($\eta$), and that a
non-anchor error is an insertion rather than a substitution ($\rho$)
(Appendix~\ref{app:channel}); $A$ is an alignment of $\hat{Y}_i$ to $Y^{c}_i$, and
$E_j(A)$, $D_j(A)$ mark that $\hat{y}_j$ is an error and that words are deleted before
it under $A$ (\S\ref{sec:supervision}). $t_j$ is the frame at which $\hat{y}_j$ is
emitted, and $s_j$ the LSTM state. In \textsc{Cusum}, $C$ is the running sum of
Eq.~\eqref{eq:cusum}, $a$ the first word of the open span ($\bot$ if none), and $P$, $p$
the peak of $C$ and its word.}
\label{alg:pardon}
\small
\begin{minipage}[t]{0.49\textwidth}
\begin{algorithmic}[1]
\Procedure{Train}{$\mathcal{D} = \{(\hat{X}_i, Y_i, X^{c}_i)\}_{i=1}^{N}$, $\mathcal{M}_{\theta}$}
  \ForAll{$i = 1, \ldots, N$}
    \State $\hat{Y}_i, Y^{c}_i \gets$ word decodes of $\hat{X}_i, X^{c}_i$ by $\mathcal{M}_{\theta}$
    \State $A^{0}_i \gets$ hard alignment (longest common runs)
  \EndFor
  \State $\Phi \gets$ event rates under $\{A^{0}_i\}$, e.g.\
         $\rho \gets \frac{\#\text{ins}}{\#\text{ins} + \#\text{sub}}$
  \For{10 iterations} \Comment{EM on the training pairs}
    \State \textbf{E:} $n_x \gets \sum_i \mathbb{E}_{P(A \mid Y^{c}_i, \hat{Y}_i; \Phi)}[\#x \text{ in } A]$
    \Statex \hspace{\algorithmicindent}\hspace{\algorithmicindent}by forward--backward, for each event $x$:
    \Statex \hspace{\algorithmicindent}\hspace{\algorithmicindent}error or correct word, deletion continued
    \Statex \hspace{\algorithmicindent}\hspace{\algorithmicindent}or ended, anchor or not, insertion or substitution

    \State \textbf{M:} $\pi \gets \frac{n_{\text{err}}}{n_{\text{err}} + n_{\text{cor}}}$,\;
           $\gamma \gets \frac{n_{\text{cont}}}{n_{\text{cont}} + n_{\text{end}}}$,
    \Statex \hspace{\algorithmicindent}\hspace{\algorithmicindent}\phantom{\textbf{M:}}
       $\eta \gets \frac{n_{\text{anc}}}{n_{\text{anc}} + n_{\text{non-anc}}}$,\;
       $\rho \gets \frac{n_{\text{ins}}}{n_{\text{ins}} + n_{\text{sub}}}$
  \EndFor
  \State freeze $\Phi$
  \ForAll{$i$ and tokens $j$ of $\hat{Y}_i$}
    \State $\tilde{e}_{ij} \gets \sum_{A} P(A \mid Y^{c}_i, \hat{Y}_i)\, E_j(A)$
    \State $\tilde{d}_{ij} \gets \sum_{A} P(A \mid Y^{c}_i, \hat{Y}_i)\, D_j(A) / \tilde{e}_{ij}$
    \State $\tilde{\ell}_{ij} \gets$ number of words deleted before $\hat{y}_j$ in $A^{0}_i$
    \State $\vf_{ij} \gets \Call{Features}{j}$
  \EndFor
  \State $w, w_d \gets$ ratios of negative to positive labels
  \State train $p_\theta$ on $\{(\vf_{ij}, \tilde{e}_{ij}, \tilde{d}_{ij}, \tilde{\ell}_{ij})\}$
         with Eq.~\eqref{eq:wbce}
  \State Platt-scale $e_j$ on dev, and $d_j$ on its error tokens
\EndProcedure
\Statex
\Procedure{Features}{$j$} \Comment{frames $t \le t_j$ only}
  \State $q[\cdot, j], r[\cdot, j] \gets$ banded forward--backward
  \State $\tilde{r}[t, j] \gets r[t, j] / \sum_{t'} r[t', j]$
  \State pool $h^{(13)}, h^{(24)}, v$, joint scalars under $q[\cdot, j], \tilde{r}[\cdot, j]$
  \State \Return the pooled features and $g[j]$
\EndProcedure
\algstore{pardon}
\end{algorithmic}
\end{minipage}\hfill
\begin{minipage}[t]{0.48\textwidth}
\begin{algorithmic}[1]
\algrestore{pardon}
\Procedure{Infer}{stream $\hat{X}$; $c_R, c_I, W$}
  \State $\hat{Y}' \gets [\,]$;\; $C \gets 0$;\; $a \gets \bot$;\;
         $s_0 \gets \mathbf{0}$
  \ForAll{tokens $\hat{y}_j$ emitted by $\mathcal{M}_{\theta}$, at frame $t_j$}
    \State $(e_j, d_j, \ell_j, s_j) \gets p_\theta(\Call{Features}{j}, s_{j-1})$
    \State $m_j \gets e_j\,[(1 - d_j) + d_j\,\ell_j]$ \Comment{Eq.~\eqref{eq:mj}}
    \If{$\hat{y}_j$ completes word $k$}
      \State append word $k$ to $\hat{Y}'$
      \State \Call{Cusum}{$k$, $\max_{\hat{y}_i \in \text{word } k} m_i$}
    \EndIf
  \EndFor
  \State \Call{Cusum}{$N{+}1$, $m_{N+1}$} \Comment{end of sequence}
  \State \textbf{if} $a \neq \bot$ and $P \ge c_I$ \textbf{then}
         \Call{Recourse}{$a..p$}
  \State \Return $\hat{Y}'$
\EndProcedure
\Statex
\Procedure{Cusum}{$k$, $\bar{m}_k$}
  \If{$C = 0$ and $\bar{m}_k > c_R$}
    \State $a \gets k$;\; $P \gets 0$ \Comment{span opens}
  \EndIf
  \State $C \gets \max(0,\, C + \bar{m}_k - c_R)$ \Comment{Eq.~\eqref{eq:cusum}}
  \State \textbf{if} $a \neq \bot$ and $C > P$ \textbf{then}
         $P \gets C$;\; $p \gets k$
  \If{$a \neq \bot$ and ($C = 0$ or $C \le P - c_I$)}
    \State \textbf{if} $P \ge c_I$ \textbf{then} \Call{Recourse}{$a..p$}
    \State $a \gets \bot$;\; $C \gets 0$
  \EndIf
\EndProcedure
\Statex
\Procedure{Recourse}{span $S$}
  \If{$|S| > W$} \Comment{length cap}
    \State $S \gets$ its $W$ consecutive words with max $\sum \bar{m}_k$
  \EndIf
  \State $o_L, o_R \gets$ anchor words on either side of $S$ in $\hat{Y}'$
  \State ask $Q(o_L, o_R)$; record the reply audio $X^{r}$
  \State $Y^{r} \gets$ decode of $X^{r}$ by $\mathcal{M}_{\theta}$
  \State $Z \gets \mathrm{LLM}(\hat{Y}' \text{ with } S \text{ marked}, Q, Y^{r})$
  \State replace the words of $S$ in $\hat{Y}'$ by $Z$
\EndProcedure
\end{algorithmic}
\end{minipage}
\end{algorithm*}

\section{Data Preparation}
\label{app:data}

\subsection{Corpora}

The AMI Meeting Corpus~\citep{carletta2005ami} contains about 100 hours of four-person
meetings recorded by headset microphones and far-field arrays.
NOTSOFAR-1~\citep{vinnikov2024notsofar} contains office meetings recorded in many
different conference rooms by several commercial devices at once, with close-talking
microphones worn by the participants; we keep one device per meeting, so that no speech
appears twice. CHiME-6~\citep{watanabe2020chime6} contains four-person dinner parties
recorded in private homes by distant microphone arrays and by binaural microphones worn
by each speaker. DiPCo~\citep{vansegbroeck2020dipco} contains ten dinner-party sessions
of four participants, recorded by five distant seven-microphone devices and by a
close-talking microphone per speaker. In all four, the noisy input is the first channel
of the first far-field array. Speaking turns in the two dinner-party corpora are much
shorter than in the meeting corpora.

The corpora are cut into utterances in two ways: AMI has a time stamp for every word, so
its long turns can be split between two words, whereas the other three corpora only time
each segment, so their long turns cannot be split and are left out of training. Across all of them, training excludes turns too short to carry
reliable labels (3\,s for the meeting corpora and 1.5\,s for the dinner parties), while
development and evaluation keep every turn. Table~\ref{tab:data-stats} gives the
resulting sizes.

\subsection{AMI}

\paragraph{Segments.}
Speaker turns are taken from the \texttt{only\_words} references of the AMI diarization
setup of \citet{landini2022bayesian}, which are derived from the AMI manual annotations
(version 1.6.2): every annotated word is speech, consecutive words of the same speaker
are merged into one turn, and turns are broken wherever the speaker pauses. The setup's
meeting lists follow the full-corpus-ASR partition (136 training, 18 development and 16
evaluation meetings). Two meetings in the training set (IS1003b, IS1007d) do not have far-field recording in the corpus and are excluded. Each speaker's global identifier is mapped to their headset channel
using the corpus metadata (\texttt{meetings.xml}).

\paragraph{Length criteria.}
Turns longer than 15\,s are split into pieces of
at most 15\,s, cutting only between two consecutive words of the speaker, at the midpoint
of the pause between them. For each piece, the cut is placed at the widest pause whose
midpoint lies between 7.5\,s and 15\,s from the start of the piece, falling back to
pauses closer to the start, and avoiding cuts that would leave a short remainder; if no
word boundary is available, the piece is cut at 15\,s. In the training set, turns
and pieces shorter than 3\,s are discarded: 6.37\% of turns are split, and 34.78\% of
training utterances come from split turns. The 3\,s threshold removes 66\% of turns but
only 16\% of the speech, and merging consecutive turns of the same speaker across pauses
of up to 1\,s recovers less than 1\% of the speech, so we do not merge turns. The development and evaluation sets keep turns of every length; in them 4.29\% and
6.67\% of turns are split, giving 9.33\% and 13.96\% of their utterances.

\paragraph{Reference transcripts.}
The reference of each utterance consists of the words of the speaker's manual
transcript whose start time falls inside it. Words without a time stamp are included
when the nearest time-stamped words on both sides are included. Non-word annotations
(vocal sounds, gaps and disfluency markers) and tokens consisting only of punctuation are
excluded.

\paragraph{Audio.}
The far-field and headset recordings are cut at the same boundaries, down-mixed to mono
and resampled to 16\,kHz.

\subsection{NOTSOFAR-1, CHiME-6 and DiPCo}

\paragraph{Sources.}
The three corpora are converted with the data-generation tools of the CHiME-8 distant
ASR challenge~\citep{cornell2024chime8}, which provide segment-level transcripts with a
start time, an end time, a speaker and the words of each segment. For NOTSOFAR-1 we use
the training (\texttt{240825.1\_train}), development (\texttt{240825.1\_dev1}) and
evaluation (\texttt{240629.1\_eval\_small\_with\_GT}, 80 meetings) releases. Each
NOTSOFAR-1 meeting appears once per recording device; we keep the sessions of a single
device (device code \texttt{107}), present in every meeting. The far-field input is the
first channel of the first array (\texttt{U01}), and the clean recording is the
speaker's close-talking channel.

\paragraph{Segments.}
Consecutive segments of the same speaker are joined into one turn when the pause between
them is at most 0.5\,s, no other speaker's segment overlaps the pause, no segment removed
under the criteria below lies inside it, and the joined turn is at most 15\,s long. The
transcripts have no word-level time stamps, so a turn cannot be split at a word boundary
as in AMI; in the training set, turns longer than 15\,s are therefore discarded, as are
turns shorter than 3\,s (NOTSOFAR-1) or 1.5\,s (CHiME-6, DiPCo). The development and
evaluation sets keep every turn, with turns outside these limits kept whole.

\paragraph{Reference transcripts.}
Annotation tags (laughter, noise, and similar) are removed from the reference. Segments
whose reference marks words as inaudible, unintelligible or redacted are excluded from
training, since the recording contains words that their reference lacks; they are kept,
with the tag removed, in the development and evaluation sets, where they make up a
substantial share of DiPCo. Segments with no words after tag removal are excluded from training; in development
and evaluation they are kept, and enter the evaluation whenever their close-talking
decode is not empty.

\paragraph{Cross-talk.}
The close-talking microphones of these corpora, and the binaural microphones of CHiME-6
in particular, also pick up nearby speakers. Words of another speaker that appear in the
close-talking decode are then counted as words the far-field decode deleted, which adds
noise to the labels of overlapping turns.

\subsection{Speech Synthesis}
\label{app:tts}
The synthetic clean rendering of a training utterance is its reference transcript, read
by Kokoro-82M~\citep{kokoro2025}. Punctuation-only tokens and the hyphens of partial
words are removed, and fillers are kept. Each speaker has a fixed voice, drawn from five
female and five male voices. In AMI, the voice matches the speaker's gender, which the
speaker identifier encodes; NOTSOFAR-1, CHiME-6 and DiPCo do not annotate gender, and
their speakers get a voice from the pooled list. The audio is resampled to 16\,kHz,
padded with 0.25\,s of silence on each side, and mixed with an excerpt of the free-sound
noise of MUSAN~\citep{snyder2015musan} at an SNR drawn uniformly from 20 to 30\,dB. The
synthesized replies of Appendix~\ref{app:spans} are rendered in the same way from the
reference words of the span. Every rendering is seeded by the utterance, so it is
reproducible.

\begin{table}[t]
\centering
\footnotesize
\setlength{\tabcolsep}{2pt}
\begin{tabular}{lrrrrrr}
\toprule
 & \multicolumn{2}{c}{Train} & \multicolumn{2}{c}{Dev} & \multicolumn{2}{c}{Eval} \\
\cmidrule(lr){2-3}\cmidrule(lr){4-5}\cmidrule(lr){6-7}
Corpus & Utt. & Hours & Utt. & Hours & Utt. & Hours \\
\midrule
AMI        & 27{,}861 & 62.9 & 9{,}145 & 8.8 & 8{,}128 & 8.5 \\
NOTSOFAR-1 & 3{,}533 & 6.2 & 9{,}493 & 5.5 & 16{,}557 & 11.0 \\
CHiME-6    & 30{,}778 & 28.5 & 6{,}926 & 6.5 & 9{,}940 & 5.9 \\
DiPCo      & 526 & 0.7 & 2{,}246 & 2.0 & 3{,}279 & 3.2 \\
\bottomrule
\end{tabular}
\caption{Utterances and hours of speech per set after segmentation. Training
excludes short and partly unintelligible turns; development and evaluation keep every
turn.}
\label{tab:data-stats}
\end{table}

\section{Recourse Details}
\label{app:spans}

\paragraph{Span boundaries.}
Each word $k$ is scored by $\bar{m}_k$, the highest $m_j$ among its tokens; tokens
without a score (punctuation) are ignored. Spans are found by the CUSUM of
\S\ref{sec:spans} (Algorithm~\ref{alg:pardon}): a span starts at the word where $C$
rises above zero and ends at the word where $C$ peaked, and the interruption is raised at
the first word at which $C$ has fallen $c_I$ below that peak. Spans are made of whole
words, so a request never quotes part of a word. A span longer than $W$ words is cut to
the $W$ consecutive words with the largest total score ($W = 6$ unless stated
otherwise). The two words before the span form the left anchor and the two words after
it the right anchor.

\paragraph{Start and end of the utterance.}
Near either edge of the utterance, a span may have fewer than two words on that side, so
its anchor is cut short: the left anchor never extends before the first word and the
right anchor never past the last. A span that reaches an edge has no anchor on that side
and uses the corresponding one-sided template below. At the end of the utterance, the
end-of-sequence position $\hat{y}_{N+1}$, which carries the deletions after the last
word, is scored as one more word, so trailing deletions can also open or extend a span.
A span still open at that point does not wait for $C$ to fall: it triggers an
interruption if its peak gain $P$ is at least $c_I$ and is dropped otherwise. We take the end of the utterance from the reference segmentation; a deployed system would need a
turn-end detector instead.

\paragraph{Request templates.}
The template depends on which anchors exist, with $L$ the (up to) two words before
the span, $R$ the (up to) two words after it, and $S$ the words of the span in the
noisy decode:
\begin{itemize}
  \item both anchors: \emph{``I didn't catch what you said between `$L$' and `$R$'.
  Could you repeat it?''}
  \item left anchor only: \emph{``What did you say after `$L$'?''}
  \item right anchor only: \emph{``What did you say before `$R$'?''}
  \item no anchor, as when the span covers a whole short utterance:
  \emph{``Did you say `$S$'?''}
\end{itemize}

\paragraph{Replies.}
\label{sec:simulation}
The speaker's reply is simulated in two ways: the speaker's close-talking recording of
the span, cut at its word times and widened by 400\,ms on each side, or a synthesized
rendering of its reference words (Appendix~\ref{app:tts}). Either reply is decoded on its
own by the same recognizer, starting from an empty predictor state; an empty reference
gives an empty reply.

\paragraph{Correction.}
Each span is corrected by Gemini 3.8 Flash~\citep{gemini38flash} at temperature 0, as
described in \S\ref{sec:clarification}. The prompt, and how the corrector's output
replaces the words of the span, are given in Appendix~\ref{app:prompt-corrector}.

\section{Prompts}
\label{app:prompts}
   \newcommand{\ph}[1]{$\langle$\textit{#1}$\rangle$}
   \newenvironment{promptbox}[1]
     {\par\smallskip\noindent\textbf{#1}\par\nopagebreak\smallskip\noindent
      \begin{lrbox}{0}\begin{minipage}{\dimexpr\columnwidth-2\fboxsep-2\fboxrule\relax}
      \footnotesize\raggedright\setlength{\parskip}{3pt}}
     {\end{minipage}\end{lrbox}\fbox{\usebox{0}}\par\smallskip}
Placeholders are shown in angle brackets. All calls use temperature 0.

\subsection{Corrector}
\label{app:prompt-corrector}
The corrector is called once per span, in the order in which the interruptions are
raised. Its user message contains the transcript decoded up to the interruption, with
earlier spans already replaced by their corrections and the current span marked
\texttt{[?\,\ldots\,?]}, followed by the request and the reply. Only the text inside
\texttt{<fix></fix>} is used, and it replaces the words of the span; every other word of
the transcript is left as decoded. When the reply is empty, no call is made and the
span's words are removed.

\begin{promptbox}{System}
You correct a live speech-recognition transcript of a meeting.
The recognizer flagged a stretch it probably got wrong; it is marked [? like this ?].
The system asked the speaker to repeat that stretch, and the speaker's answer is given.
The answer was also transcribed automatically, so it can contain recognition errors,
and the speaker may have repeated a few words from just before or after the stretch.

Write the corrected text for the marked stretch only:\\
- base it on the answer where the answer is plausible in context;\\
- leave out words of the answer that belong to the transcript outside the marks;\\
- keep the transcript's style (case, punctuation);\\
- if the marked words were not actually spoken, return nothing.\\
Reply with the corrected stretch inside \texttt{<fix></fix>} and nothing else.
\end{promptbox}

\begin{promptbox}{User}
Transcript so far:\\
\ph{words before the span} [?\ph{span words}?] \ph{words after the span, up to the interruption}

System asked: \ph{request (Appendix~\ref{app:spans})}\\
Speaker answered: ``\ph{transcribed reply}''
\end{promptbox}

\subsection{$N$-best Correction}
\label{app:prompt-nbest}
The $N$-best baseline uses the instruction template of HyPoradise~\citep{chen2023hyporadise}
in a single user turn, preceded by ten demonstrations in the same format. The best
hypothesis is the greedy transcript, which every other method corrects; the other
hypotheses are the four highest-scoring beam hypotheses that remain distinct from it
and from each other after text normalization. The demonstrations are the longest
development-set utterances of at most 40 words, after removing utterances whose greedy
hypothesis is far from the reference (cross-talk in the reference). The answer is read
after \texttt{\#\#\# Response:}; an empty or overly long answer is replaced by the greedy
transcript. The $N$-best oracle picks, for each utterance, the hypothesis of the same list
closest to the reference.

\begin{promptbox}{System}
You correct speech recognition transcripts. Follow the format of the examples exactly
and reply with the corrected transcription only.
\end{promptbox}

\begin{promptbox}{User}
Below is a best-hypotheses that is transcribed from an automatic speech recognition
system. Write a response to predict the true transcription using the tokens from
other-hypotheses.

Example 1:\\
\#\#\# Best-hypothesis:\\
\ph{greedy transcript}

\#\#\# Other-hypothesis:\\
\ph{hypothesis 2}\\
\ldots\\
\ph{hypothesis 5}

\#\#\# Response:\\
\ph{reference}

\ldots\ (Examples 2--10)

Now the utterance to correct:\\
\#\#\# Best-hypothesis:\\
\ph{greedy transcript}

\#\#\# Other-hypothesis:\\
\ph{hypothesis 2}\\
\ldots\\
\ph{hypothesis 5}

\#\#\# Response:
\end{promptbox}

\section{Implementation of the Learned Confidence Estimators}
\label{app:token-baselines}
\label{app:peng}
All four learned estimators of \S\ref{sec:baselines} read the same frozen transducer and
are trained on the same corpus, partition and labels as the CF-Gain Scorer: the hard
labels $E_j$ from the close-talking decode, with early stopping on development-set
average precision. Every subword token is scored, and a word error marks all of its
tokens. Each estimator is made causal: token $j$, emitted at frame $t_j$, is scored only
from the tokens up to $j$ and the encoder frames up to $t_j$, as the CF-Gain Scorer is.

\paragraph{\citet{qiu2021multitask}.}
Each token's input is its embedding and position, the decoder activation, its log
posterior and the top-$K$ log posteriors, read at its greedy emission point.
\begin{itemize}
  \item Self-attention is masked to tokens up to $j$. Cross-attention covers only the 16
  encoder frames ending at $t_j$, instead of the whole utterance.
  \item The deletion head predicts our deletion anchor, the token after the gap, as an
  auxiliary task. The utterance-confidence head is an auxiliary task during training
  only. The token score is the output of the main head.
  \item We omit the multi-hypothesis input, which needs complete beams.
\end{itemize}

\paragraph{\citet{wang2021rnnt}.}
\begin{itemize}
  \item The three frames around the emission, $t_j-1, t_j, t_j+1$, are replaced by the
  three frames ending at it, $t_j-2, t_j-1, t_j$, since $t_j+1$ may fall in the next
  encoder chunk.
  \item Self-attention covers the past 30 tokens only.
\end{itemize}

\paragraph{TruCLeS~\citep{ravi2025trucles}.}
\begin{itemize}
  \item The input is the same as for \citet{wang2021rnnt}, and the LSTMs are
  unidirectional instead of bidirectional.
  \item The confidence target is computed with the close-talking decode as the
  reference. The probability of the reference token is read under the noisy decode's
  prefix, since the streaming recognizer never sees the true one. The error score is
  one minus the predicted confidence.
\end{itemize}

\paragraph{\citet{peng2026proactive}.}
We reimplement the detectors from the description in the paper.
\begin{itemize}
  \item \textbf{Detectors.} We keep the perception-error detector, which scores each
  token from the input to the recognizer's joint network, and the deletion detector,
  which scores encoder frames. We omit the comprehension-error detector: it is trained
  on errors against the reference transcript, so it would learn errors that a
  re-utterance does not correct. We also omit the distortion-type detector, whose labels
  require synthetic distortion conditions and whose output is used only by their own
  dialogue policy.
  \item \textbf{Deletion targets.} The paper does not specify where on the frame axis a
  deletion target is placed. We mark the frames the noisy decode spent waiting to emit
  the deletion's anchor token, those whose dwell posterior is at least 1\% of its
  maximum (the final row of the lattice for a deletion after the last token). Deletion
  scores are pooled to a token over the blank frames that precede it, matching our
  anchoring convention.
  \item \textbf{Combination.} The scores of the two detectors are each mapped to
  percentiles fitted on the development set and combined by taking the maximum.
\end{itemize}
\section{Pair-HMM Channel}
\label{app:channel}

\begin{figure}[t]
  \centering
  \begin{tikzpicture}[x=0.98cm, y=1cm, >={Stealth[length=1.6mm]},
  tok/.style={font=\small\strut, inner sep=1.5pt},
  lab/.style={font=\scriptsize, text=black!60},
  num/.style={font=\small}]
  \definecolor{exA}{RGB}{15,110,86}
  \definecolor{exB}{RGB}{83,74,183}
  \definecolor{err}{RGB}{186,64,40}
  \node[lab, anchor=east] at (-0.35,2) {clean};
  \foreach \w/\x/\n in {i/0/c1, think/1/c2, that/2/c3, that/3/c4, is/4/c5, fine/5/c6}
    \node[tok] (\n) at (\x,2) {\w};
  \node[lab, anchor=east] at (-0.35,0.6) {noisy};
  \foreach \w/\x/\n in {i/0/n1, think/1/n2, that/2.5/n3, is/4/n4, fine/5/n5}
    \node[tok] (\n) at (\x,0.6) {\w};
  \foreach \a/\b in {c1/n1, c2/n2, c5/n4, c6/n5}
    \draw[black!35] (\a.south) -- (\b.north);
  \draw[exA, thick] (c3.south) -- (n3.north);
  \draw[exB, thick, dashed] (c4.south) -- (n3.north);
  \node[lab, anchor=east] at (-0.35,-0.15) {hard};
  \node[lab, anchor=east] at (-0.35,-0.7) {soft $\tilde{e}_j$};
  \foreach \x/\h/\q in {0/0/0, 1/0/0, 2.5/0/0.5, 4/1/0.5, 5/0/0} {
    \node[num] at (\x,-0.15) {\ifnum\h=1 \textcolor{err}{\h}\else \h\fi};
    \node[num] at (\x,-0.7) {\q};
  }
  \draw[decorate, decoration={brace, mirror, amplitude=4pt}, thick]
    (2.2,-1.0) -- (4.3,-1.0) node[midway, below=4pt, font=\scriptsize] {a 2-token span covers both};
\end{tikzpicture}
  \caption{An ambiguous deletion. Two alignments explain the noisy decode equally well:
  the second \emph{that} was deleted and \emph{is} anchors the deletion (solid), or the
  first was deleted and the surviving \emph{that} anchors it (dashed). The hard alignment
  labels only one token and treats the other as clean; the soft label splits the mass
  between them, so neither is penalized, and a span covering both carries the same label
  mass either way.}
  \label{fig:soft}
\end{figure}

\begin{figure}[t]
  \centering
  \begin{tikzpicture}[x=0.74cm, y=0.7cm, >={Stealth[length=1.5mm]},
  st/.style={draw, rounded corners=2pt, minimum width=0.62cm, minimum height=0.46cm,
             font=\footnotesize, inner sep=1.5pt},
  jn/.style={circle, draw, fill=black!8, inner sep=1.2pt},
  lab/.style={font=\scriptsize, fill=white, inner sep=0.6pt},
  note/.style={font=\scriptsize, text=black!55}]
  \definecolor{ok}{RGB}{15,110,86}
  \definecolor{err}{RGB}{186,64,40}
  \node[st] (s) at (0,0) {$\hat{y}_j$};
  \node[st, draw=ok, text=ok] (C) at (4.2,2.1) {C};
  \node[jn] (E) at (1.3,-1.2) {};
  \node[jn] (e0) at (4.0,0.2) {};
  \node[st, draw=err, text=err] (D) at (2.6,-3.0) {D};
  \node[jn] (e1) at (4.0,-3.0) {};
  \node[st, draw=err, text=err] (I) at (7.4,0.9) {I};
  \node[st, draw=err, text=err] (S) at (7.4,-1.0) {S};
  \node[st, draw=err, text=err] (A) at (7.4,-3.0) {A};
  \node[st] (n) at (9.0,2.1) {$\hat{y}_{j+1}$};

  \draw[->] (s) -- node[lab, above left] {$1-\pi$} (C);
  \draw[->] (s) -- node[lab, left] {$\pi$} (E);
  \draw[->] (E) -- node[lab, above left, pos=0.5] {$1-\gamma$} (e0);
  \draw[->] (E) -- node[lab, left] {$\gamma$} (D);
  \draw[->] (D) edge[loop below] node[lab, below] {$\gamma$} (D);
  \draw[->] (D) -- node[lab, above] {$1-\gamma$} (e1);
  \draw[->] (e0) -- node[lab, above, pos=0.5] {$\rho$} (I);
  \draw[->] (e0) -- node[lab, above right, pos=0.45] {$1-\rho$} (S);
  \draw[->] (e1) -- node[lab, left, pos=0.3] {$(1-\eta)\rho$} (I);
  \draw[->] (e1) -- node[lab, below right, pos=0.55] {$(1-\eta)(1-\rho)$} (S);
  \draw[->] (e1) -- node[lab, below, pos=0.5] {$\eta$} (A);
  \draw[->] (C) -- (n);
  \draw[->] (I.east) -- ++(0.5,0) |- (n);
  \draw[->] (S.east) -- ++(0.5,0) |- (n);
  \draw[->] (A.east) -- ++(0.5,0) |- (n);
  \node[note, anchor=south] at ($(e0.north)+(0,1pt)$) {$m=0$};
  \node[note, anchor=north] at ($(e1.south)+(0,-2pt)$) {$m\geq1$};
\end{tikzpicture}
  \caption{The pair-HMM channel at noisy token $\hat{y}_j$. A correct token (C) is written
  unchanged. An erroneous one is preceded by a run of deleted tokens (D, possibly empty)
  and is then an anchor (A, written unchanged; only after a deletion), an insertion
  (I, nothing written) or a substitution (S, a different token written). C has $E_j = 0$;
  A, I and S have $E_j = 1$.}
  \label{fig:channel}
\end{figure}
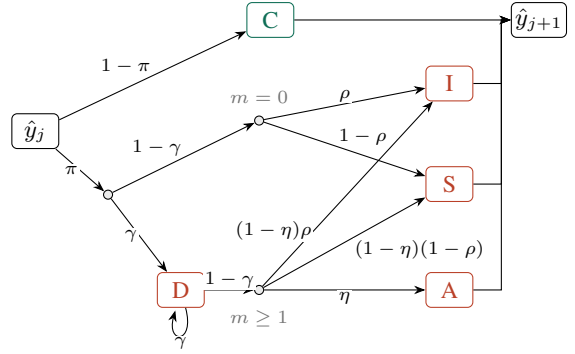

The soft labels of \S\ref{sec:supervision} come from a pair hidden Markov
model~\citep{durbin1998biological} that reads the noisy decode $\hat{Y}$ token by token
and writes the clean decode $Y^{c}$ (Figure~\ref{fig:channel}). Figure~\ref{fig:soft}
shows the case it is built for: a deletion that two alignments explain equally well.

Each noisy token carries a latent label $E_j \in \{0, 1\}$, and four parameters govern
the channel: the probability that a token is in error ($\pi$), that a run of deleted
tokens continues ($\gamma$), that an erroneous token is the anchor of a preceding
deletion ($\eta$), and that a non-anchor error is an insertion rather than a
substitution ($\rho$). They are fit by expectation maximization on the training pairs,
initialized from the counts of the hard alignment, and frozen after ten iterations
(Algorithm~\ref{alg:pardon}). The forward--backward pass of the fitted model gives the
posterior $\tilde{e}_j = P(E_j = 1 \mid Y^{c}, \hat{Y})$ for every noisy token, including
the end-of-sequence position, and in the same way the posterior that $\hat{y}_j$ is
preceded by deleted tokens (the D state of Figure~\ref{fig:channel}); dividing it by
$\tilde{e}_j$ gives $\tilde{d}_j$. The number of deleted words $\tilde{\ell}_j$ is read from the hard
alignment. Wherever the two decodes determine the alignment, $\tilde{e}_j$ equals the hard
label $E_j$. The channel runs on normalized words, and its labels are mapped to subword
tokens as described in \S\ref{app:implementation}.

\newcommand{\recrow}[2]{\makebox[4.6em][l]{\textit{#1}}%
  \parbox[t]{\dimexpr\columnwidth-4.6em\relax}{\raggedright #2}\par\vspace{1pt}}
\newenvironment{recex}[1]
  {\par\noindent\begin{minipage}{\columnwidth}\footnotesize\textbf{#1}\par\vspace{2pt}}
  {\end{minipage}\par\vspace{8pt}}

\section{Recourse Examples}
\label{app:rec-examples}

\begin{recex}{Substitution repaired}
  \recrow{Decode}{Yeah, \textbf{[we get behind schedule yeah]} pretty quick like you say do we want to set a date for interim prototype?}
  \recrow{Request}{I didn't catch what you said between `Yeah' and `pretty quick'. Could you repeat it?}
  \recrow{Reply}{It'll get behind schedule, yeah.}
  \recrow{Corrected}{Yeah, it'll get behind schedule yeah pretty quick like you say do we want to set a date for interim prototype?}
  \recrow{Reference}{Yeah it'll get behind schedule yeah pretty quick Like you're saying yeah Well actually do we wanna set a date for interim prototype}
\end{recex}

\begin{recex}{Substitution repaired}
  \recrow{Decode}{\textbf{[We have]} the number buttons, I guess.}
  \recrow{Request}{What did you say before `the number'?}
  \recrow{Reply}{Leave out}
  \recrow{Corrected}{Leave out the number buttons, I guess.}
  \recrow{Reference}{leave out the number buttons I guess}
\end{recex}

\begin{recex}{Deleted words restored}
  \recrow{Decode}{Okay, is that how it works in the other ones in like the other windows like various like um \textbf{[yeah topic]} or}
  \recrow{Request}{I didn't catch what you said between `like um' and `or'. Could you repeat it?}
  \recrow{Reply}{If we're in the topic}
  \recrow{Corrected}{Okay, is that how it works in the other ones in like the other windows like various like um if we're in the topic or}
  \recrow{Reference}{Okay Is that how it works in the other ones In like the other windows like the various like um if we're in the topic Or}
\end{recex}

\begin{recex}{Deleted words restored}
  \recrow{Decode}{\textbf{[This is not really]} like for the group you were making this all really require a lot of marketing.}
  \recrow{Request}{What did you say before `like for'?}
  \recrow{Reply}{It's pretty different, but this is not really}
  \recrow{Corrected}{It's pretty different, but this is not really like for the group you were making are remote, this will really require a lot of marketing.}
  \recrow{Reference}{It's pretty different But this is not really like for the group we were making our remote This will really require a lot of marketing}
\end{recex}

\begin{recex}{Inserted words removed}
  \recrow{Decode}{Um, \textbf{[actually this]} first thing I just sort of did a really sort of brief um yeah version of what we have to do}
  \recrow{Request}{I didn't catch what you said between `Um' and `first thing'. Could you repeat it?}
  \recrow{Reply}{Actually}
  \recrow{Corrected}{Um, actually first thing I just sort of did a really sort of brief um yeah version of what we have to do}
  \recrow{Reference}{Um actually first thing I just sorta did a really sorta brief um yeah version of what we have to do Um actually I can just take that if you want}
\end{recex}

\begin{recex}{Partly repaired}
  \recrow{Decode}{Nice and pink fluorescent banana color or whatever you know \textbf{[make a banana chip on you]} for example, you know, but the question is really who is our target group?}
  \recrow{Request}{I didn't catch what you said between `you know' and `for example'. Could you repeat it?}
  \recrow{Reply}{Yeah.}
  \recrow{Corrected}{Nice and pink fluorescent banana color or whatever you know for example, you know, but the question is really who is our target group?}
  \recrow{Reference}{nice and pink fluorescent banana colour or whatever you know yeah for example you know But the question is really who is our target group}
\end{recex}

\begin{recex}{Partly repaired}
  \recrow{Decode}{\textbf{[Is]} there anything you would like to to ask me about the findings before I?}
  \recrow{Request}{What did you say before `there anything'?}
  \recrow{Reply}{I know. Okay, well I hope that's a clear to the team is}
  \recrow{Corrected}{I know. Okay, well I hope that's clear to the team. Is there anything you would like to to ask me about the findings before I?}
  \recrow{Reference}{I know Okay Well I hope that's uh clear to the team Is there anything you would like to to ask me about the findings before I}
\end{recex}

\begin{recex}{False alarm, transcript unchanged}
  \recrow{Decode}{Maybe we should just have two, but yeah, if \textbf{[he finds them then that's cool]} if he finds the people um.}
  \recrow{Request}{I didn't catch what you said between `yeah, if' and `if he'. Could you repeat it?}
  \recrow{Reply}{He finds them then that's cool.}
  \recrow{Corrected}{Maybe we should just have two, but yeah, if he finds them then that's cool if he finds the people um.}
  \recrow{Reference}{yeah Three or four Oh God Um but yeah if he finds them then that's cool If he finds the people M um}
\end{recex}

\begin{recex}{False alarm, transcript unchanged}
  \recrow{Decode}{\textbf{[All these changes are not really]} okay with me, but since}
  \recrow{Request}{What did you say before `okay with'?}
  \recrow{Reply}{All these changes are not really.}
  \recrow{Corrected}{All these changes are not really okay with me, but since}
  \recrow{Reference}{all these changes are not really okay with me But since}
\end{recex}

\begin{recex}{False alarm, error introduced}
  \recrow{Decode}{Um like we have the the right click menu \textbf{[menu]} right clicking but instead of like have a default double click so there's so it'll be a choice}
  \recrow{Request}{I didn't catch what you said between `click menu' and `right clicking'. Could you repeat it?}
  \recrow{Reply}{Menu.}
  \recrow{Corrected}{Um like we have the the right click menu right clicking but instead of like have a default double click so there's so it'll be a choice from the menu}
  \recrow{Reference}{Um like we have the the right-click menu You'll have the menu right-click it but instead of like have a default double-click so if there's a de so it'll be a choice from the menu but yeah exactly Yeah}
\end{recex}

\begin{recex}{False alarm, error introduced}
  \recrow{Decode}{\textbf{[There something like that,]} but that's that sounds a lot cheaper to me.}
  \recrow{Request}{What did you say before 'but that's'?}
  \recrow{Reply}{There's something like that.}
  \recrow{Corrected}{There's something like that, but that's that sounds a lot cheaper to me.}
  \recrow{Reference}{there something like that but that's that sounds a lot cheaper to me}
\end{recex}

\begin{recex}{Failure: the reply was misrecognized}
  \recrow{Decode}{Yeah \textbf{[text yeah yeah yeah cause that's]} the like declares a variable isn't and then yeah because}
  \recrow{Request}{I didn't catch what you said between `Yeah' and `the like'. Could you repeat it?}
  \recrow{Reply}{Yeah yeah yeah cause that}
  \recrow{Corrected}{Yeah yeah yeah cause that the like declares a variable isn't and then yeah because}
  \recrow{Reference}{Yeah Yeah yeah 'cause that's the uh the declares a variable doesn't it and then they keep Okay}
\end{recex}

\begin{recex}{Failure: the reply was misrecognized}
  \recrow{Decode}{\textbf{[The whistle]} so we can financially voice recognition is not feasible we could go for a whistle.}
  \recrow{Request}{What did you say before `so we'?}
  \recrow{Reply}{For hell yeah, the whistle.}
  \recrow{Corrected}{The whistle so we can financially voice recognition is not feasible we could go for a whistle.}
  \recrow{Reference}{For l yeah the whistle So we can if financially voice recognition is not feasible we could go for a whistle}
\end{recex}

\begin{recex}{Failure: correct reply, wrong correction}
  \recrow{Decode}{Right \textbf{[attributes you've got lots of nom]} something and then night something and then end something}
  \recrow{Request}{I didn't catch what you said between `Right' and `something and'. Could you repeat it?}
  \recrow{Reply}{Yeah, you've got lots of Nom}
  \recrow{Corrected}{Right you've got lots of nom something and then night something and then end something}
  \recrow{Reference}{Oh yeah You've got lots of nom something and then NITE something and then N\_ something}
\end{recex}

\begin{recex}{Failure: correct reply, wrong correction}
  \recrow{Decode}{\textbf{[Very so we've]} got twenty cents for our docking station.}
  \recrow{Request}{What did you say before `got twenty'?}
  \recrow{Reply}{Very subtitles buttons yeah seven yeah so we've}
  \recrow{Corrected}{So we've got twenty cents for our docking station.}
  \recrow{Reference}{Very Subtitles buttons Yeah Seven Yeah So we've got twenty cents for our docking station huh}
\end{recex}
\section{Out-of-Domain Generalization with Matched Training Size}
\label{app:ood-subset}
The source corpora of Table~\ref{tab:ood} have far more training data than their targets
(62.9 vs.\ 6.2 hours and 28.5 vs.\ 0.7 hours; Table~\ref{tab:data-stats}). To separate
the effect of the corpus from that of the training size, we train a CF-Gain Scorer on a
subset of the source's training set with as many hours as the target's training set.
The subset is built from whole sessions: the sessions are visited in a fixed random
order and their utterances are added until the target duration is reached, so the
subset holds a few complete sessions and part of one, like a smaller corpus rather than
a random sample of turns. This gives 11 complete AMI meetings and part of a twelfth, out of 134 training meetings, and part of one of the 16 CHiME-6 training sessions. The
scorer is otherwise trained as in \S\ref{sec:implementation}, with model selection and
calibration on the source's development set, and applied unchanged to the target.

Table~\ref{tab:ood-subset} shows the results. With matched training size, the
transferred scorer falls below the in-domain scorer on both pairs: on NOTSOFAR-1 by 4.0
points of AP and 11.1 points of recall at 70\% precision, and on DiPCo by 2.3 points of
AP and 6.7 points of recall. The gains of Table~\ref{tab:ood} therefore come from the
larger training set of the source corpus: at equal size, in-domain data is worth more,
but a larger corpus of the same genre can replace it. The CHiME-6 subset comes from a
single session, so its result likely depends on which session is drawn.

\begin{table}[t]
\centering
\small
\setlength{\tabcolsep}{3.5pt}
\resizebox{\columnwidth}{!}{%
\begin{tabular}{llrccc}
\toprule
 & & & & \multicolumn{2}{c}{Recall at prec.} \\
\cmidrule(lr){5-6}
Test & Trained on & Hours & AP & 50 & 70 \\
\midrule
\multirow{4}{*}{NOTSOFAR-1} & Entropy         & --   & 39.2 & 28.6 & 6.3 \\
                            & NOTSOFAR-1      & 6.2  & 59.8 & 68.7 & 39.4 \\
                            & AMI             & 62.9 & \textbf{63.2} & \textbf{71.7} & \textbf{39.7} \\
                            & AMI, subset     & 6.2  & 55.8 & 60.9 & 28.3 \\
\midrule
\multirow{4}{*}{DiPCo}      & Entropy         & --   & 41.3 & 29.0 & 11.0 \\
                            & DiPCo           & 0.7  & 44.1 & 36.6 & 10.0 \\
                            & CHiME-6         & 28.5 & \textbf{54.8} & \textbf{58.0} & \textbf{24.4} \\
                            & CHiME-6, subset & 0.7  & 41.8 & 31.4 & 3.3 \\
\bottomrule
\end{tabular}}
\caption{Out-of-domain generalization with matched training size. A CF-Gain Scorer
trained on the other corpus of the same genre, on all of its training set or on a subset
with as many hours as the target's training set, is applied unchanged to the target and
compared with a scorer trained on the target and with the training-free entropy baseline.
All numbers except hours are in percent. Best result per block in bold.}
\label{tab:ood-subset}
\end{table}
\section{Speech Enhancement Baseline}
\label{app:enhancement}
The far-field recording is passed through DeepFilterNet3~\citep{schroeter2023deepfilternet},
a causal noise-suppression model with about 40\,ms of algorithmic latency that suits
a streaming recognizer, and the enhanced audio is decoded by the same recognizer. The
model runs at 48\,kHz; the 16\,kHz input is resampled up and the output resampled back.
We cap its attenuation at a limit fitted on the development set. Enhancement increases
the WER on both evaluation sets, from 30.0\% to 32.4\% on AMI (relative WER reduction
$-8.0\%$) and from 64.1\% to 67.0\% on CHiME-6 ($-4.5\%$), so it is not shown in
Figure~\ref{fig:recourse}.
\section{Additional Recourse Results}
\label{app:recourse-more}
\begin{figure*}[t]
  \centering
  \begin{minipage}[t]{0.48\textwidth}
    \centering
    \includegraphics[width=\linewidth]{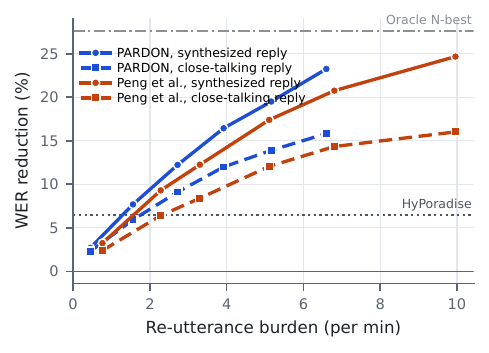}\\
    {\small (a) AMI, $\beta = 0.10$, at most six words}
  \end{minipage}\hfill
  \begin{minipage}[t]{0.48\textwidth}
    \centering
    \includegraphics[width=\linewidth]{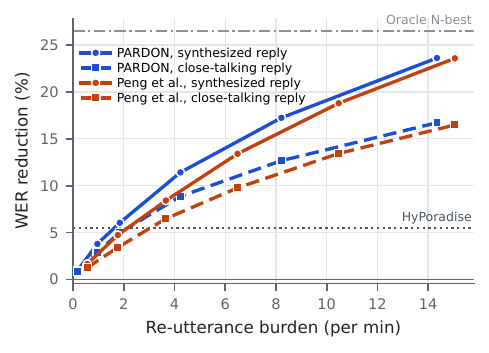}\\
    {\small (b) CHiME-6, $\beta = 0.10$, at most six words}
  \end{minipage}\\[0.8em]
  \begin{minipage}[t]{0.48\textwidth}
    \centering
    \includegraphics[width=\linewidth]{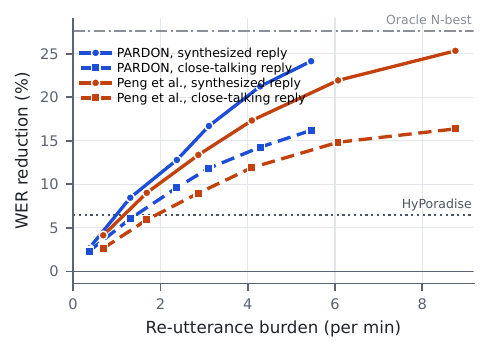}\\
    {\small (c) AMI, $\beta = 0.05$, at most eight words}
  \end{minipage}\hfill
  \begin{minipage}[t]{0.48\textwidth}
    \centering
    \includegraphics[width=\linewidth]{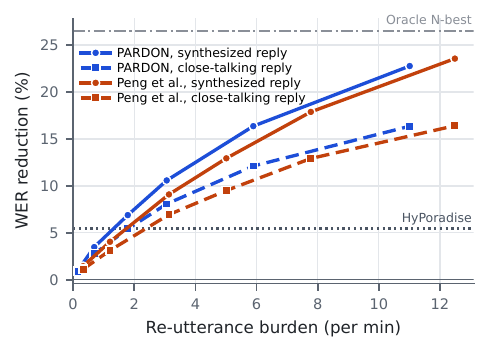}\\
    {\small (d) CHiME-6, $\beta = 0.05$, at most eight words}
  \end{minipage}
  \caption{Recourse on the AMI and CHiME-6 evaluation sets with a higher weight on
  re-uttered words (a, b) and with longer spans (c, d): relative WER reduction against
  the re-utterance burden, at the operating points selected on the development set,
  with synthesized or close-talking replies. Horizontal lines: the zero-interruption
  $N$-best correction of HyPoradise and the $N$-best oracle.}
  \label{fig:recourse-more}
\end{figure*}

Figure~\ref{fig:recourse-more} repeats the recourse comparison of
\S\ref{sec:recourse-results} with a higher weight on re-uttered words ($\beta = 0.10$,
spans of at most six words) and with longer spans (at most eight words, $\beta = 0.05$).
Operating points are selected on the development set as in \S\ref{app:costtuning}.

\section{Tuning Cost Parameters}
\label{app:costtuning}
We find operating points on the development set by minimizing the burden for different
budgets of recall, the share of erroneous words, counting each deleted word, that fall
inside a span or in the gap before one of its words. Recall approximates the ceiling on
the error reduction our system can achieve. It is measured against the close-talking
decode, so no gold transcript is used: a substituted or inserted word counts when it lies
inside a span, and the deleted words of a gap count when the gap lies inside a span or at
its boundary. Utterances with an empty noisy decode stay in the denominator.

Operating points are chosen separately for each scorer, span cap $W$ and weight $\beta$.
We sweep $c_R \in \{0.3, 0.4, \ldots, 0.8\}$ and 64 values of $c_I$ spaced geometrically
from 0.02 to 20. The recall budgets are six values spaced evenly from 5\% to 90\% of the
lower of the two scorers' highest development recall, so that both scorers can reach
every budget. For each budget, the pair $(c_R, c_I)$ with the lowest development burden
among those that reach it is applied unchanged to the evaluation set. The
zero-interruption baselines are reported as single points.
\section{Training Details}
\label{app:implementation}

\paragraph{Supervision.}
The close-talking recording is transcribed by the same ASR system as the far-field
audio, and this transcript serves as the clean decode for supervision. Both noisy and
clean decodes are normalized to words (lower-cased, with punctuation removed) before a
pair HMM aligns them. Word-level substitution and insertion labels are copied to every
subword token of the noisy word, and deletion labels, together with $\tilde{d}_j$ and
$\tilde{\ell}_j$, only to its first token.
The pair HMM is initialized from the hard-alignment counts, fit with ten EM
iterations on the training utterances, and frozen before the CF-Gain Scorer is trained.

\paragraph{Features.}
Encoder states are pooled from the 13th and the final (24th) encoder blocks under the
emission and dwell posteriors,
giving four 1024-dimensional vectors per token, and are concatenated with the
predictor state and the scalar features of \S\ref{sec:features}. Voice activity
probabilities come from Silero VAD~\citep{silero2021vad}.

\paragraph{Training.}
The CF-Gain Scorer is a single-layer unidirectional LSTM with 128 hidden units, followed
by three heads, each a feed-forward network with one hidden layer of 64 units. It is
trained with AdamW~\citep{loshchilov2019adamw} at a learning rate of $10^{-3}$ and weight
decay $10^{-4}$, on batches of 64 utterances, for up to 60 epochs with early stopping on
development-set AP of $e_j$ (patience 15). We then apply Platt scaling to $e_j$ on the
development set, and to $d_j$ on its error tokens. A separate scorer is trained on each corpus.
\paragraph{Compute.}
All experiments run on a single NVIDIA L4 GPU. On AMI, the largest corpus, extracting
the lattice features of the training set takes about 4 hours and training the CF-Gain
Scorer about 30 minutes; all experiments in this paper, including the baselines,
ablations, speech synthesis and recourse evaluation, took about 40 GPU hours in total.
The LLM corrector and the $N$-best baseline are queried through the Gemini API. All
results are from a single training run per setting.
\end{document}